\documentclass[10pt,twocolumn,letterpaper]{article}

\usepackage[dvipdfmx]{graphicx}
\usepackage{subfigure}
\usepackage{siunitx}

\usepackage[pagenumbers]{iccv} 

\definecolor{iccvblue}{rgb}{0.21,0.49,0.74}
\usepackage[pagebackref,breaklinks,colorlinks,allcolors=iccvblue]{hyperref}

\def\paperID{10800} 
\def\confName{ICCV}
\def\confYear{2025}

\title{Rethink 3D Object Detection from Physical World}

\author{Satoshi Tanaka, \quad Koji Minoda, \quad Fumiya Watanabe, \quad Takamasa Horibe\\
TIER IV, Inc\\
{\tt\small satoshi.tanaka@tier4.jp}
}

\begin{document}
\maketitle
\begin{abstract}

High-accuracy and low-latency 3D object detection is essential for autonomous driving systems.
While previous studies on 3D object detection often evaluate performance based on mean average precision (mAP) and latency, they typically fail to address the trade-off between speed and accuracy, such as 60.0 mAP at \SI{100}{ms} vs 61.0 mAP at \SI{500}{ms}.
A quantitative assessment of the trade-offs between different hardware devices and accelerators remains unexplored, despite being critical for real-time applications.
Furthermore, they overlook the impact on collision avoidance in motion planning, for example, 60.0 mAP leading to safer motion planning or 61.0 mAP leading to high-risk motion planning.
In this paper, we introduce latency-aware AP (L-AP) and planning-aware AP (P-AP) as new metrics, which consider the physical world such as the concept of time and physical constraints, offering a more comprehensive evaluation for real-time 3D object detection.
We demonstrate the effectiveness of our metrics for the entire autonomous driving system using nuPlan dataset, and evaluate 3D object detection models accounting for hardware differences and accelerators.
We also develop a state-of-the-art performance model for real-time 3D object detection through latency-aware hyperparameter optimization (L-HPO) using our metrics.
Additionally, we quantitatively demonstrate that the assumption "the more point clouds, the better the recognition performance" is incorrect for real-time applications and optimize both hardware and model selection using our metrics.

\end{abstract}

\section{Introduction}
\label{sec:intro}

\begin{figure*}[t]
    \begin{center}
        \includegraphics[width=0.96\linewidth]{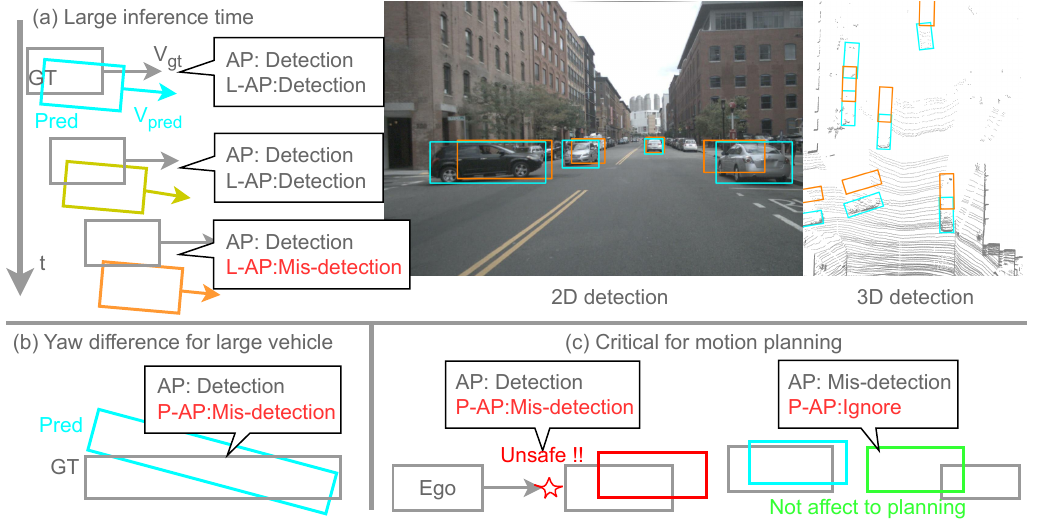}
        \caption{
            The cases where a high score on existing mAP does not correspond to practical performance.
            Grey objects represent ground truths (GTs), while color objects represent prediction (Pred) results.
            The image of (a) shows inference latency for 2D and 3D object detection.
            Blue objects represent the results with a latency of \SI{0}{ms}, while orange objects represent the results with a latency of \SI{500}{ms}.
        }
        \label{concept}
    \end{center}
\end{figure*}

As autonomous driving technology advances, the ability to perceive and understand the environment is crucial.
While 2D object detection has been a mainstay in computer vision, 3D object detection is vital for autonomous vehicles to navigate complex environments safely.
3D object detection is more challenging than 2D object detection because it requires understanding both position and orientation with more dimensions, which is critical for tasks such as collision avoidance in motion planning.
Research in 3D object detection has focused on the average precision (AP) metric, with state-of-the-art models being evaluated based on mean average precision (mAP) using open datasets such as nuScenes \cite{nuscenes} and Waymo Open Dataset \cite{Waymo2020}.

However, when considering real-time applications in robotics, such as autonomous vehicles, traditional evaluation metrics like mAP may not accurately reflect real-world performance.
As shown in \cref{concept}, a model might achieve a high mAP yet still perform poorly in practice.
\cref{concept} (a) shows the case where detection accuracy is high, but inference time is too long.
This is primarily because 3D object detection metrics have been adapted from those used in 2D object detection, which do not account for the dynamics of robotic systems.
Evaluation in computer vision is typically conducted in a static environment, whereas robotics operates in a time-dependent context.
As shown in \cref{concept}, while the error in image space is small even with \SI{500}{ms} latency and IoU (Intersection over Union) is sufficient in 2D object detection, the results show discrepancies, as the IoU in the BEV (Bird’s Eye View) space approaches 0 in 3D object detection.
Moreover, \cref{concept} (b) shows the model may treat yaw measurements for large and small vehicles identically, even though larger vehicles’ yaw fluctuations have a greater impact.
Additionally, \cref{concept} (c) shows failure to detect critical objects for motion planning can lead to unsafe conditions, highlighting the gap between traditional mAP metrics and real-world safety requirements.

For issue of latency, while previous works on 3D object detection often evaluate performance based on mean average precision (mAP) and latency, they typically fail to address the trade-off between speed and accuracy, such as \rm{60.0} mAP at \SI{100}{ms} vs \rm{61.0} mAP at \SI{500}{ms}.
A quantitative assessment of the trade-offs between device variations and accelerators has yet to be researched, despite being critical for real-time applications.
Moreover, even in cases where performance improves using large data inputs, such as merge of multi-frame LiDAR pointcloud, a trade-off between performance and inference time exists, which has yet to be quantitatively assessed.
In practical applications, while using more powerful GPUs or employing accelerators like TensorRT to reduce latency can sometimes "improve recognition performance," evaluations have thus far been limited to assessments along the latency axis only.
In addition to latency, they leave the impact on collision avoidance in motion planning, for example, \rm{60.0} mAP leading to safer motion planning or \rm{61.0} mAP leading to high-risk motion planning.
Even if the mAP is high, a model becomes unusable if it fails to detect objects that are critical for motion planning.

To address these issues, we rethink about the metrics for 3D object detection in the context of the physical world, especially "time" and "spatial perception".
It becomes evident that current evaluation methods in 3D object detection do not account for "time," a fundamental concept in physics with a long research history.
Moreover, current evaluation methods do not account for "spatial perception", a fundamental for motion planning which need the information to avoid collision for planning-aware objects.
In this study, we proposes metrics for 3D object detection in real-time applications, by incorporating the concept of physicality.
The primary contribution of this work is to bridge the gap between academic algorithms in robotics perception and those suitable for real-world implementation, advancing the development of robotics perception.
Specifically, we make three key contributions:

\begin{enumerate}
\item We introduce latency-aware AP (L-AP) and planning-aware AP (P-AP) as new evaluation metrics, and demonstrate their effectiveness as a metrics for motion planning in real-world scenarios using nuPlan dataset.
\item We evaluate current baseline 3D object detection models accounting for device differences and the use of accelerators using our metrics, and develop a high-performance model for real-time 3D object detection that surpasses existing models by performing latency-based hyperparameter optimization.
\item We show the optimal amount of point cloud in real-time 3D object detection and the optimization for hardware and model selections through quantitative evaluation.
\end{enumerate}

\section{Related Work}
\label{sec:background}

\textbf{Real-time 3D object detection.}
3D object detection \cite{Chen20203DPC, mao20233dobjectdetectionautonomous} is mainly divided into LiDAR-based and camera-based approaches.
Researchers have developed 3D object detection models using LiDAR point cloud \cite{Qi2016PointNetDL, Qi2017, Zhou2017VoxelNetEL, Yan2018SECONDSE, Yang2018PIXORR3, Shi2018PointRCNN3O,Shi2019PVRCNNPF, Yang20203DSSDP3}.
PointPillars \cite{Lang2018PointPillarsFE} stands out as a model with performance and speed, representing a breakthrough for LiDAR-based 3D object detection in real-time robotics.
CenterPoint \cite{yin2021center} simplifies the architecture by using a center-based representation, making it a popular baseline and practical choice in industry applications.
There has also been research into 3D object detection models based on camera-LiDAR fusion \cite{Chen2016Multiview3O, Qi2018, Vora2019PointPaintingSF, wang2021pointaugmenting, Drews2022DeepFusionAR, Yingwei2022, Huang2022MultimodalSF, Li2022DelvingIT, Liang2022BEVFusionAS, Huang2024DL, li2024fully}.
TransFusion \cite{Bai2022TransFusionRL} is a baseline model using an attention-based method.
BEVFusion \cite{liu2022bevfusion} unifies bird’s-eye view (BEV) representation from multiple sensors and improves the performance.
Camera-based 3D object detection also has been developed in computer vision researches \cite{Ali2018YOLO3DER, Tian2019FCOSFC}.
LSS \cite{Philion2020} is commonly used as a baseline method in camera-based 3D object detection, where it lifts 2D image features to a BEV grid in 3D space.
Recently, multi-camera 3D object detection models have gained attention \cite{Huang2021BEVDetHM,detr3d, Yang2022BEVFormerVA, Li2022BEVDepthAO, Zhang2022BEVerseUP, Wang2023ExploringOT, JiangLLWJWHZ24}.
BEVFormer \cite{li2022bevformer} represents a breakthrough in multi-camera 3D object detection using the transformer mechanism and BEV representation, which transforms local image features from a 2D image encoder into BEV space.
Several studies have focused on improving real-time performance in 3D object detection.
\cite{Wang2022CostAwareEA} proposes an evaluation method for LiDAR-based 3D object detection that considers computational costs and adjusts the model accordingly.
LidarNAS \cite{Liu2022} uses neural architecture search (NAS) to enhance real-time 3D object detection.
UTR3D \cite{Chen2022FUTR3DAU} aims to achieve real-time 3D object detection with a cost-effective sensor kit.

\textbf{Metrics for 3D object detection.}
As metrics for 3D object detection, the most common evaluation metric is Average Precision (AP), which extends the 2D mAP metric by assessing object detection accuracy in 3D space, accounting for both spatial overlap and object localization across all three dimensions.
nuScenes detection score (NDS), proposed in the nuScenes dataset \cite{nuscenes}, combines various performance metrics including AP, orientation, and velocity to evaluate detection performance.
In Waymo Open Dataset \cite{Waymo2020}, not only AP but also AP weighted by heading (APH) is used for 3D object detection.

In addition to the major metrics proposed from datasets, several metrics have been reported to improve detection performance.
Stability index \cite{WangMLYWCH24} provides the metrics of consistency of 3D object detection models to create stable 3D object detection model.
Pl-metrics \cite{Ivanovic2021InjectingPI} introduce a planning cost function into perception evaluation and analyze the relationship between perception failures and motion planning.
OMNI3D \cite{brazil2023omni3d} raises concerns about the adequacy of IoU as the sole evaluation metric for 3D object detection performance.
LET-3D-AP \cite{Hung2024} introduces a modified 3D AP that accounts for longitudinal errors, improving the evaluation of camera-only 3D object detection.
Long-range Detection Score (LDS) \cite{Yang2024} focuses on long-range 3D object detection.
MultiCorrupt \cite{Till2024} evaluates the robustness of LiDAR-camera fusion models for 3D object detection, with an emphasis on diverse sensor data corruptions.
Object Criticality Model \cite{Ceccarelli2023} shows the importance of assessing object detection systems not only for accuracy but also for their safety and reliability.
MTBF model \cite{Oboril2022MTBFMF} aims to construct safety evaluation considering perception error.

\cite{Ivanovic2021InjectingPI, Ceccarelli2023, Oboril2022MTBFMF} analyze the relationship between task-driven metrics, however, they do not focus on developing higher-performance models while considering latency.
\cite{WangMLYWCH24, brazil2023omni3d, Hung2024, Yang2024, Till2024} apply their proposed metrics to high-performance models but do not examine the relationship between these metrics and downstream tasks.
Furthermore, to the best of our knowledge, no metric has been reported that incorporates latency and enables single-axis evaluation.

\section{Method}
\label{sec:method}

\subsection{Latency-aware Average Precision (L-AP)}
\label{sec:method-1}

\begin{figure}[t]
    \begin{center}
        \includegraphics[width=0.95\linewidth]{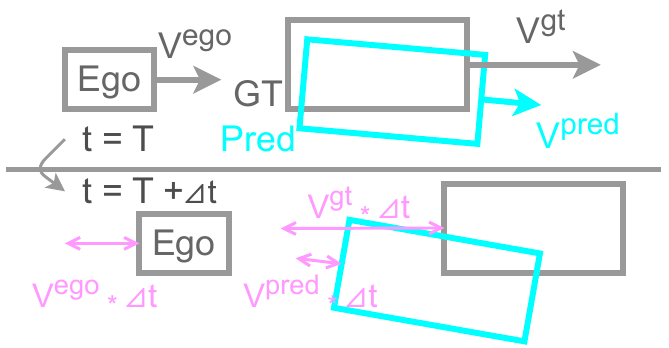}
        \caption{
          The metric of L-AP.
          L-AP measures the accuracy of detection at the moment when the inference is completed, considering the latency involved.
        }
        \label{mAP-latency}
    \end{center}
\end{figure}

\begin{figure*}[t]
    \begin{minipage}[b]{0.38\linewidth}
        \centering
        \includegraphics[width=0.95\linewidth]{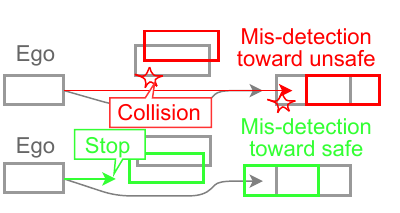}
        \hspace{1.6cm} (a)
    \end{minipage}
    \begin{minipage}[b]{0.38\linewidth}
        \centering
        \includegraphics[width=0.97\linewidth]{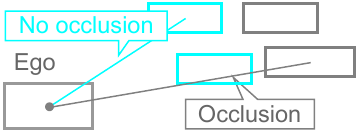}
        \hspace{1.6cm} (b)
    \end{minipage}
    \begin{minipage}[b]{0.23\linewidth}
        \centering
        \includegraphics[width=0.97\linewidth]{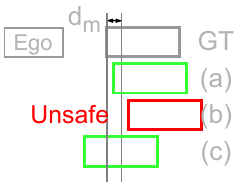}
        \hspace{1.6cm} (c)
    \end{minipage}
    \caption{
      (a) Mis-detection towards unsafe and safe.
      If a 3D object detection model misidentifies objects as being farther away than their actual positions, it can lead to planning errors, increasing the risk of collisions by causing inaccurate collision avoidance.
      In contrast, misidentifying objects as being closer than their actual positions in 3D object detection generally results in a lower risk of collisions in planning, showing behavior that appears to stop earlier.
      (b) Occlusion filtering for planning-aware objects.
      Occluded objects do not significantly impact for motion planning in many cases.
      (c) Matching algorithm in planning-aware objects.
      $ d_m $ is the margin parameter used in planning.
      If the distance error exceeds $ d_m $ and the object is farther than the ground truth, the object is considered not to match.
    }
    \label{planning-aware}
\end{figure*}

\begin{table}[t]
  \centering
  \caption{
    Evaluation results with P-mAP for \cref{planning-aware} (c).
    $ d_e $ represents the error in the distance to the nearest surface.
    We use distance thresholds for P-mAP as (0.5, 1.0, 1.5, 2.0) [m].
  }
  \begin{tabular}{|c|c|c|c|} \hline
  & (a) & (b) & (c) \\ \hline
  $ d_e $ & +0.25 & +0.75 & -0.75 \\ \hline
  Is matching? & \textcolor{blue}{Yes} & \textcolor{red}{No} & \textcolor{blue}{Yes} \\ \hline
  P-mAP & 1.0 & 0.0 & 0.75 \\ \hline
  \end{tabular}
  \label{planning-aware-3}
\end{table}

\begin{figure}[t]
    \begin{center}
        \includegraphics[width=0.95\linewidth]{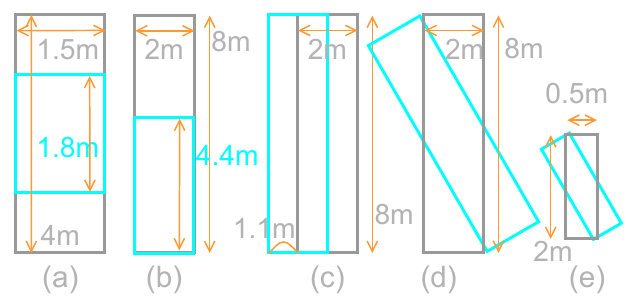}
        \caption{
          The case to evaluate of corner distance.
          Grey objects represent ground truths, while blue objects represent prediction results.
          (a) Varying vehicle sizes.
          (b) Different vertical positions of large vehicles.
          (c) Different horizontal positions of large vehicles.
          (d) Yaw deviation of $ \pi/6 $ in a large vehicle.
          (e) Yaw deviation of $ \pi/6 $ in a bicycle.
        }
        \label{corner_distance}
    \end{center}
\end{figure}

\begin{table}[t]
  \centering
  \caption{
    Comparison of different metrics across various cases in \cref{corner_distance}.
    We use distance thresholds for mAP as (0.5, 1.0, 1.5, 2.0) [m].
    IoU is 0.55 in (b) and 0.27 in (d) and we use IoU thresholds for mAP as (0.3, 0.5).
  }
  \begin{tabular}{|l|c|c|c|c|c|} \hline
    & (a) & (b) & (c) & (d) & (e) \\ \hline
    mAP@IoU & 0.5 & 1.0 & 0.5 & 0.0 & 0.0 \\
    Planning-aware? & \textcolor{blue}{Yes} & \textcolor{red}{No} & \textcolor{blue}{Yes} & \textcolor{blue}{Yes} & \textcolor{red}{No} \\ \hline
    mAP@Center & 1.0 & 0.25 & 0.5 & 1.0 & 1.0 \\
    Planning-aware? & \textcolor{red}{No} & \textcolor{blue}{Yes} & \textcolor{blue}{Yes} & \textcolor{red}{No} & \textcolor{blue}{Yes} \\ \hline
    mAHS@Center & 1.0 & 0.25 & 0.5 & 0.83 & 0.83 \\
    Planning-aware? & \textcolor{red}{No} & \textcolor{blue}{Yes} & \textcolor{blue}{Yes} & \textcolor{red}{No} & \textcolor{blue}{Yes} \\ \hline
    mAP@Corner & 0.75 & 0.25 & 0.5 & 0 & 0.75 \\
    Planning-aware? & \textcolor{blue}{Yes} & \textcolor{blue}{Yes} & \textcolor{blue}{Yes} & \textcolor{blue}{Yes} & \textcolor{blue}{Yes} \\ \hline
  \end{tabular}
  \label{metrics_comparison}
\end{table}

In this work, we introduce Latency-aware Average Precision (L-AP) in the evaluation of 3D object detection, with the goal of integrating the concept of "time" into real-time 3D object detection.
As shown in \cref{mAP-latency}, we measure how accurate the detection is at the moment the inference is completed.
Specifically, we evaluate true positive (TP) metrics by adjusting the evaluation based on the velocity multiplied by latency, which effectively shifts the detection results in time.
To begin, $ \hat{v_{t}}^{GT} $, the velocity of the ground truth object at time $ t $ is calculated by

\begin{equation}
    \hat{v_{t}}^{GT} = \frac{x_{t}^{GT} - x_{t - \delta t}^{GT}}{\delta t},
\end{equation}

where  $ x_{t}^{GT} $ represents the position of the object in ground truth at time $ t $, and $ \delta t $ is the time interval between annotations.
Next, we estimate the position of the object at time $ (t + \Delta t) $ by

\begin{equation}
    x_{t + \Delta t}^{GT} = x_{t}^{GT} + \hat{v_{t}}^{GT} \cdot \Delta t,
\end{equation}

where $ \Delta t $ is the inference time.
Note that we will discuss the accuracy of $x_{t + \Delta t}^{GT}$ in Appendix \ref{sec:appendix-2}.
The predicted position at time $ (t + \Delta t) $ is calculated by

\begin{equation}
    x_{t + \Delta t}^{Pred} = x_{t}^{Pred} + v_{t}^{Pred} \cdot \Delta t - v_{t}^{Ego} \cdot \Delta t,
\end{equation}

where $ x_{t}^{Pred} $ represents the position of the predicted object at time $ t $, $ v_{t}^{Pred} $ is the predicted velocity from the result of 3D object detection at time $ t $, and $ v_{t}^{Ego} $ is the velocity of the ego vehicle at time $ t $.

By incorporating latency into the evaluation, we provide a more practical metric for real-world applications of 3D object detection.
Specifically, accounting for latency allows for a more accurate assessment of detection performance in operational settings.
First, latency highlights that evaluation results can vary depending on the deployment hardware.
For instance, upgrading the hardware can improve detection performance, as reduced latency leads to faster inference.
However, traditional AP metrics do not directly consider latency, meaning these hardware improvements may not be fully captured in conventional evaluations.
Furthermore, introducing latency enables a more consistent assessment of the benefits provided by acceleration libraries, such as TensorRT.
These libraries are designed to optimize inference speed, and by incorporating latency into the evaluation, we can more uniformly measure their impact on overall detection performance.

\subsection{Planning-aware Average Precision(P-AP)}
\label{sec:method-2}

In the context of perception systems, mis-detection can be categorized into two distinct types: safe mis-detection and unsafe mis-detection as shown in \cref{planning-aware} (a).
Safe mis-detections are those where the system's errors do not lead to hazardous consequences.
On the other hand, unsafe mis-detections are the ones which have the potential to cause serious errors, such as collisions or failure to perform critical tasks.
Typically, false negatives or overestimates of the distance from an ego-vehicle to a target vehicle turn out to be critical for planning algorithms.
While traditional evaluation metrics often fail to differentiate between these types of mis-detections, it is critical for autonomous systems, particularly in real-time applications, to properly account for unsafe mis-detections, as they pose significant safety risks.

Therefore, we propose a metric of Planning-aware Average Precision (P-AP).
First, we apply occlusion filtering to focus on planning-aware objects, as shown in \cref{planning-aware}(b).
Objects that are not occluded and have a significant impact on motion planning are considered planning-aware objects and included in the evaluation.
Additionally, P-AP regards mis-detections occurring at farther distances as overestimation, reflecting the increased risk posed by such errors, as illustrated in \cref{planning-aware}(c).
While mis-detections closer to the object tend to result in the system stopping safely at the front, those occurring at greater distances can lead to dangerous collisions.
In robotics, it is common practice to plan trajectories with a margin, which is often set as a parameter.
In this work, we define the margin parameter $ d_m $ for planning as and set it to 0.5 [m].
In our proposed matching process, we use the distance to the nearest surface instead of the center position, which is often used in the matching step in AP calculation, because motion planning should focus on the surface closest to the ego vehicle rather than the center position.
If the error of the distance of the nearest surface $ d_e $ exceeds $ d_m $ and the object is further than ground truth, the object is considered unsafe and not matching to ground truth.
We calculate the scores with the penalty of farther distance in cases like those illustrated in \cref{planning-aware} (c) and summarized in \Cref{planning-aware-3}.

In addition to filtering and matching for planning-aware objects, we replace matching metrics from center distance to corner distance.
This modification is motivated by principles from motion planning.
In motion planning, the primary concern is not overlap-based metrics such as Intersection over Union (IoU) or the center position but rather the safety margin required to ensure a collision-free trajectory.
Specifically, the focus is on determining the necessary clearance to avoid collisions, rather than merely assessing object proximity in the detection space.
In practice, the center-based evaluation used in AP for nuScenes does not penalize minor misalignments, such as diagonal displacements, making it suboptimal for planning purposes.
Moreover, evaluation metrics that account for yaw errors, such as AHS used in Waymo Open Dataset \cite{Waymo2020}, excessively penalizes from yaw errors for small objects such as pedestrians and bicycles in the same way as for trailers.

Instead of center distance, we calculate the average displacement of the four corner points.
First, the corner distance for each corner point $d_c(i)$ is calculated by $ d_c(i) = \sqrt{(x_i^{\text{Pred}} - x_i^{\text{GT}})^2} \quad (i \in P) $, where $i$ represents each corner point, $ P $ is set of corner points of the bounding box, $ x_i^{\text{Pred}} $ is the position of the predicted object's corner, and $ x_i^{\text{GT}}$ is the position of the ground truth object's corner.
The overall corner distance $d_c$ is then computed as the average of the distances across all four corners by $ d_c = \frac{1}{4} \sum_{i \in P} d_c(i) $.

In \cref{corner_distance}, we compare the performance of different metrics across various real-world cases, including mAP(IoU), mAP(Center distance), mAHS, and P-mAP as shown in \Cref{metrics_comparison}.
For the cases in \cref{corner_distance} (a) and (d), P-mAP is able to handle mis-detections of size, while AP and AHS fail to account for this factor.
In the cases of \cref{corner_distance} (b) and (e), P-mAP does not overestimate or underestimate the value, showing a reasonable decrease.
However, AP(IoU) tends to produce a high score when size detection fails, or it underestimates the detection of small objects that are not significantly misaligned in absolute terms.

\subsection{Latency-aware Hyperparameter Optimization}
\label{sec:method-3}

Building on our previous proposals, we further conduct latency-aware hyperparameter optimization (L-HPO) to enhance real-time 3D object detection.
By incorporating latency into the optimization process, we can achieve optimal performance while ensuring timely decision-making and maintaining system reliability in applications like robotics and autonomous vehicles.
Existing methods typically optimize by assigning weights to both latency and metrics such as mAP.
However, by using the proposed metric, optimization can be performed directly along an evaluation axis, enabling a more efficient search for the optimal model.
In this paper, we use the CenterPoint \cite{yin2021center} model as base model to optimize hyperparameters such as the number of layers, the number of channels in each layer, the number of multi-frame to merge point clouds, and the voxel size, aiming to find the model for the highest L-mAP.

\section{Experiment}
\label{sec:exp}

\subsection{Preparation}
\label{sec:exp-1}

In \cref{sec:exp-2}, to evaluate the effectiveness of our metrics as a comprehensive performance indicator for the application based on 3D object detection, we evaluate the overall behavior of an autonomous driving system using nuScenes dataset \cite{nuscenes} and nuPlan dataset \cite{Karnchanachari2024TowardsLP}.
For 3D object detection, we use ground truth as prediction results and added errors in latency of inference, yaw degree of objects, and the position of objects to them in nuPlan dataset.
To evaluate the contribution of the metrics to motion planning, we use PDM-Hybrid \cite{Dauner2023CORL} as the baseline with nuPlan dataset.
In our experiments, we use the closed-loop metric as nuPlan score used in \cite{Dauner2023CORL}.

From \cref{sec:exp-3} to \cref{sec:exp-5}, we evaluate 3D object detection based on our metrics using the validation dataset of the nuScenes dataset.
The hardware devices used for testing by an RTX3090 and an RTX4060Ti.
The software backend includes PyTorch and TensorRT \cite{TensorRT}, an optimized deep learning inference library by NVIDIA that accelerates latency reduction on GPUs.
For the 3D object detection models, we use PointPillars \cite{Lang2018PointPillarsFE}, CenterPoint\cite{yin2021center}, TransFusion-L \cite{Bai2022TransFusionRL} (LiDAR-only model), BEVFusion \cite{liu2022bevfusion}, and BEVFormer \cite{li2022bevformer} (base model).
LiDAR-only model of TransFusion is prepared for comparison with BEVFusion.
The inference time for reference is based on the respective papers and their TensorRT implementation \cite{NVIDIAAIIOT}.

\subsection{Evaluation for Metrics}
\label{sec:exp-2}

\begin{table*}[t]
    \caption{
      Evaluation in L-mAP and P-mAP with latency, yaw difference, and position difference using nuPlan dataset.
      NuPlan refers to the score based on the closed-loop metric for nuPlan dataset.
      (a) Verification of effectiveness in L-mAP for latency.
      "Lat." refers to latency.
      (b) Verification of effectiveness in P-mAP for yaw difference.
      "Yaw" refers to the yaw difference from the ground truth.
      (c) Verification of effectiveness in P-mAP for the position error of planning-aware object.
      "Difference" refers to the positional difference between the predicted and ground truth object positions.
      A positive value means the predicted objects are placed farther, and a negative value means the predicted objects is placed closer.
    }
    \begin{minipage}[b]{0.28\linewidth}
        \centering
        \begin{tabular}{ c | c c  }
        Lat.         & L-mAP & nuPlan \\ \hline
        \SI{0}{ms}   & 100.0 & 0.9277 \\
        \SI{50}{ms}  & 98.6  & 0.9274 \\
        \SI{100}{ms} & 94.3  & 0.9267 \\
        \SI{200}{ms} & 88.5  & 0.9143 \\
        \SI{500}{ms} & 78.2  & 0.8978 \\ \hline
        \end{tabular}
        \hspace{1.2cm} (a)
    \end{minipage}
    \begin{minipage}[b]{0.27\linewidth}
        \centering
        \begin{tabular}{ c | c c  }
            Yaw  & P-mAP & nuPlan \\ \hline
              0  & 100.0 & 0.9277 \\
        $\pi$/24 & 91.1  & 0.8539 \\
        $\pi$/12 & 78.5  & 0.7976 \\
        $\pi$/6  & 60.2  & 0.7183 \\
        $\pi$/4  & 49.7  & 0.6544 \\ \hline
        \end{tabular}
        \hspace{1.2cm} (b)
    \end{minipage}
    \begin{minipage}[b]{0.43\linewidth}
        \centering
        \begin{tabular}{ r | c l c }
        Difference & mAP & P-mAP & nuPlan \\ \hline
        \SI{0}{m}                    & 100.0 & 100.0 & 0.9277 \\
        -\SI{0.6}{m}($<$ \SI{20}{m}) &  83.2 & 76.6 (-6.6)  & 0.9354 \\
        +\SI{0.6}{m}($<$ \SI{20}{m}) &  83.3 & 24.7 (-58.6) & 0.8578 \\
        -\SI{0.6}{m}($>$ \SI{20}{m}) &  77.3 & 79.2 (+1.9)  & 0.9310 \\
        +\SI{0.6}{m}($>$ \SI{20}{m}) &  77.3 & 25.0 (-52.3) & 0.9248 \\ \hline
        \end{tabular}
        \hspace{1.2cm} (c)
    \end{minipage}
    \label{exp-metrics}
\end{table*}

\Cref{exp-metrics} (a) presents the results of L-mAP as latency of 3D object detection increases.
We observe that nuPlan score decreases in accordance with the increase in the latency, and L-mAP also correspondingly declines.
\Cref{exp-metrics} (b) presents the results of P-mAP with the error in yaw angle of objects.
We observe that nuPlan score decreases in accordance with the increase in yaw difference, and P-mAP also correspondingly declines.
Notably, in both cases, mAP remains consistently at 100.0, whereas L-mAP and P-mAP provide a more relevant evaluation metric by incorporating planning-based considerations.
These results demonstrate that L-mAP and P-mAP better reflect the actual performance of real-time 3D object detection than traditional mAP.

\Cref{exp-metrics} (c) presents the results of P-mAP with the difference in object positions.
When errors are introduced to objects close to the ego vehicle, the value of P-mAP decreases, and nuPlan score similarly decreases.
For objects close to the ego vehicle, when they are shifted slightly, mAP drops from 83.2 to P-mAP 76.6, and nuPlan score increases because the metrics related to collision risk improve.
However, when these objects are displaced farther away, mAP drops from 83.3 to P-mAP 24.7, and nuPlan score also decreases accordingly.
This quantitatively demonstrates the dangerous scenario of "mis-detection of nearby objects as distant" and shows that this metric can effectively reflect performance degradation in the planning metric.
Moreover, when objects are displaced over longer distances, the impact on nuPlan score is smaller compared to objects displaced nearby.
This behavior is well-represented by P-mAP, which aligns closely with the planning metric.

\subsection{Benchmark with Our Metrics}
\label{sec:exp-3}

\begin{table*}[t]
  \centering
  \caption{
    Evaluation results for 3D object detection on the nuScenes validation dataset using an RTX3090 GPU and the PyTorch environment.
    We recalculated mAP using the MMDetection3D base framework, establishing a baseline for comparison.
  }
  {
  \begin{tabular}{c|crccc|cc}
                    & Modality & Latency & GPU memory &NDS & mAP & L-mAP & P-mAP \\ \hline
  PointPillars \cite{Lang2018PointPillarsFE} & LiDAR & \SI{34}{ms} & 7512MiB & 52.6 & 39.0 & 38.0  & 8.4 \\
  CenterPoint\cite{yin2021center}            & LiDAR & \SI{80}{ms} & 4285MiB & 64.8 & 56.3 & 46.7  & 36.9  \\
  TransFusion-L \cite{Bai2022TransFusionRL}  & LiDAR & \SI{80}{ms} & 4412MiB & 69.1 & 64.3 & 53.6  & 27.0 \\
  BEVFusion \cite{liu2022bevfusion}  & Camera, LiDAR & \SI{119}{ms}& 8246MiB & 71.2 & 68.4 & 49.2  & 26.0 \\
  BEVFormer \cite{li2022bevformer}          & Camera & \SI{416}{ms}& 5435MiB & 51.8 & 41.7 & 18.3 & 18.4 \\ \hline
  L-HPO (Ours)                              & LiDAR  & \SI{45}{ms} & 4136MiB & 64.8 & 57.7 & \textbf{55.1} & \textbf{36.6} \\ \hline
  \end{tabular}
  }
  \label{exp-model}
\end{table*}

\begin{table}[t]
  \centering
  \caption{
    Performance comparison of different models with latency and various environments.
    (P) refers to the PyTorch backend, and (T) refers to the TensorRT backend.
    "Lat." stands for latency.
  }
  \begin{tabular}{c c | r c }
  Environment  & Model         &  Lat. & L-mAP \\ \hline
  RTX4060Ti(P) & CenterPoint   &  \SI{197}{ms} & 31.6 \\
  RTX4060Ti(P) & TransFusion-L &  \SI{202}{ms} & 35.9 \\
  RTX4060Ti(P) & BEVFusion  &  \SI{277}{ms} & 31.2 \\
  RTX4060Ti(P) & L-HPO (Ours)  &  \SI{123}{ms} & \textbf{41.0} \\\hline
  RTX3090(P)   & CenterPoint   &  \SI{80}{ms}  & 46.7 \\
  RTX3090(P)   & TransFusion-L &  \SI{80}{ms}  & 53.6 \\
  RTX3090(P)   & BEVFusion  &  \SI{119}{ms} & 49.2 \\
  RTX3090(P)   & L-HPO (Ours)  &  \SI{45}{ms}  & \textbf{55.1} \\ \hline
  RTX4060Ti(T) & CenterPoint   &  \SI{33}{ms}  & 55.0 \\
  RTX4060Ti(T) & TransFusion-L &  \SI{34}{ms}  & 63.0 \\
  RTX4060Ti(T) & BEVFusion  &  \SI{47}{ms}  & \textbf{64.8} \\
  RTX4060Ti(T) & L-HPO (Ours)  &  \SI{20}{ms}  & 57.1 \\ \hline
  RTX3090(T)   & CenterPoint   &  \SI{10}{ms}  & 56.2 \\
  RTX3090(T)   & TransFusion-L &  \SI{9}{ms}   & 64.3 \\
  RTX3090(T)   & BEVFusion  &  \SI{20}{ms}  & \textbf{68.0} \\
  RTX3090(T)   &  L-HPO (Ours) &  \SI{7}{ms}  & 57.5 \\ \hline
  \end{tabular}
  \label{exp-hardware}
\end{table}

In this study, we conducted an evaluation with our metrics to assess the real-time performance of various 3D object detection models, including our proposed model in \cref{sec:method-3}.
The results are shown in \Cref{exp-model}.
L-HPO outperforms other models in L-mAP and demonstrates high performance for real-time 3D object detection applications.
Specifically, L-HPO increase L-mAP by +8.4 from CenterPoint, by +1.5 from TransFusion-L, by +5.9 from BEVFusion.
L-HPO surpasses P-mAP by +9.6 from TransFusion-L and +10.6 from BEVFusion despite L-HPO being lightweight similar to PointPillars.
In addition to it, it is comparable performance in P-mAP to CenterPoint despite L-HPO being lighter than CenterPoint.
These results suggest that L-HPO is a better choice in the context of real-time robotics.

As for the analysis of other models, it is noteworthy that, when comparing TransFusion-L and BEVFusion in terms of L-mAP, the L-mAP of BEVFusion decreased by -4.4 (from 53.6 to 49.2) while its mAP increased by +4.1 (from 64.3 to 68.4).
This indicates that the additional latency introduced by Camera-LiDAR fusion led to a performance drop in real-time 3D object detection.
When comparing BEVFormer with PointPillars between Camera-based detection and LiDAR-based detection, mAP of BEVFormer exceed the performance of PointPillars (41.7 vs 39.0), however, BEVFormer do not exceed in L-mAP (18.3 vs 38.0), meaning that BEVFormer can be considered unsuitable for real-time applications.
CenterPoint provides highest performance in terms of P-mAP, and both TransFusion-L and BEVFusion were unable to achieve high scores in P-mAP unlike mAP, making CenterPoint a more favorable model for motion planning tasks in real-time applications.

Additionally, we evaluated the impact of weaker GPUs and acceleration libraries on performance in \Cref{exp-hardware}.
Device performance improvements, such as reducing latency, directly contribute to better perception performance in autonomous driving systems in practical scenarios.
By upgrading from RTX4060Ti to RTX3090, we observed improvements in L-mAP as +15.1 in CenterPoint, +17.7 in TransFusion-L, +18.0 in BEVFusion, and +14.1 in L-HPO.
Regarding the contribution of TensorRT acceleration, using accelerators to reduce latency improved real-time performance significantly.
TensorRT improved the following models in L-mAP as +23.4 in CenterPoint, +27.1 in TransFusion-L, +33.8 in BEVFusion, and +16.1 in L-HPO.
By using a single-axis metric for evaluation, it becomes possible to compare performance across different devices and accelerators.
For example, "RTX4060Ti + TensorRT + CenterPoint" outperforms "RTX3090 + Pytorch + TransFusion-L" in L-mAP (55.0 vs 53.6) and it suits for application with real-time 3D object detection.

\subsection{Rich Input Data Makes Better Perception?}
\label{sec:exp-4}

\begin{table}[t]
  \centering
  \caption{
    Performance evaluation of multi-frame point clouds for CenterPoint with RTX3090 GPU and Pytorch backend.
    N is the number of frames to merge point clouds.
    $P_N$ is the number of merged point clouds.
  }
  \begin{tabular}{c|r r | c c c c}
  $N$ & $P_N$     & Latency & mAP & L-mAP & P-mAP \\ \hline
  1 & 34k  & \SI{31}{ms} & 46.7 & 45.5 & 28.5 \\
  3 & 102k & \SI{44}{ms} & 52.5 & \textbf{50.3} & 35.9 \\
  6 & 204k & \SI{67}{ms} & 55.4 & 48.3 & 35.9 \\
  9 & 306k & \SI{99}{ms} & \textbf{56.3} & 43.7 &  \textbf{36.9} \\
  \hline
  \end{tabular}
  \label{exp-densification}
\end{table}

In previous studies, merging several frame point clouds as one input has primarily been used to improve the accuracy of 3D object detection models.
However, these approaches have often overlooked the real-time performance requirements critical in real-world applications.
The assumption that a higher density of LiDAR point clouds automatically leads to better recognition accuracy is a common, yet oversimplified.
This assumption fails to account for the concept of "time", which are fundamental in real-time applications.
Despite the widespread nature of this misconception, there is a noticeable lack of studies that quantitatively demonstrate its impact.
To address this gap, we evaluate the optimal number of LiDAR points required for real-time robotics.
In this study, we focus on measuring latency as the sum of preprocessing time to merge point clouds and the model inference time.
Since each frame contains approximately 34k LiDAR points in nuScenes dataset, the merged point clouds with 9 frames increases to about 300k points.
At this case, the processing time to merge multi-frame point clouds takes \SI{19}{ms}, making the total inference time 99ms with 80ms of the model's own inference time.

\Cref{exp-densification} shows the result for multi-frame point clouds.
As the number of input points increased, we observed improvements in mAP, P-mAP, indicating a positive correlation between point cloud density and detection accuracy.
However, when considering L-mAP, a different trend emerged.
Specifically, the highest L-mAP was achieved 50.3 at 3 frames, but it decreased as the frames increased: 48.3 at 6 frames and 43.7 at 9 frames.
This indicates that the increased latency due to the larger point cloud size, had a greater negative impact on the model's performance than the improvements in accuracy brought by denser point clouds.
Our findings demonstrate that, in real-time environments where low latency is critical, indiscriminately increasing the input data size offers diminishing returns.
This study provides a quantitative metric to show that a balance must be struck between accuracy and latency to achieve optimal performance in real-time 3D object detection for robotics.

\subsection{Optimize development cost with hardware}
\label{sec:exp-5}

\begin{table}[t]
    \centering
    \caption{
      Cost comparison for different models and backends with different devices.
      "Cen." represents CenterPoint and "Trans." represents TransFusion-L.
      "(P)" refers to the PyTorch backend, and "(T)" refers to the TensorRT backend.
      "C-1" represents cost for 1 systems, "C-10" for 10 systems, "C-100" for 100 systems (\$).
    }
    \begin{tabular}{cc|c|c|c|c}
        Model  & Device    & L-mAP & C-1  & C-10 & C-100 \\ \hline
        Cen.   & 4060Ti(P) & 31.6  & 21k  & 30k  & 120k \\
        Cen.   & 3090(P)   & 46.7  & 24k  & 60k  & 420k \\
        Cen.   & 4060Ti(T) & 55.0  & 41k  & 50k  & 140k \\
        Cen.   & 3090(T)   & 56.2  & 44k  & 80k  & 440k \\ \hline
        Trans. & 4060Ti(P) & 35.9  & 41k  & 50k  & 140k \\
        Trans. & 3090(P)   & 53.6  & 44k  & 80k  & 440k \\
        Trans. & 4060Ti(T) & 63.0  & 61k  & 70k  & 160k \\
        Trans. & 3090(T)   & 64.3  & 64k  & 100k & 460k \\ \hline
    \end{tabular}
    \label{exp-cost}
\end{table}

Furthermore, as an industry case study, we explore the optimization of development costs and hardware selection using L-mAP.
By using L-mAP, we enable quantitative comparisons that extend beyond differences in hardware devices and acceleration libraries.
In this study, we formulate an optimization problem where L-mAP serves as the objective function, considering both hardware and development costs.
The hardware cost is defined as \$4k for a system with an RTX3090 GPU and \$1k for a system with an RTX4060Ti GPU.
The initial development cost is defined as \$20k for implementing CenterPoint in PyTorch (three months of work for a single engineer).
Upgrading the algorithm from CenterPoint to TransFusion-L in PyTorch requires an additional cost of \$20k, while developing a TensorRT version of either CenterPoint or TransFusion-L requires an additional \$20k.
The experiment considers a business scenario in which 1, 10, or 100 perception systems are deployed, with the goal of determining the most cost-effective approach to maximizing L-mAP.

For a budget of \$60k allocated to deploying 10 perception systems, we compare three strategies: upgrading the hardware by running CenterPoint in PyTorch on an RTX3090, using an accelerator by implementing CenterPoint in TensorRT on an RTX4060Ti, and upgrading the algorithm by employing TransFusion-L in PyTorch on an RTX4060Ti.
The results indicate that upgrading the hardware leads to an L-mAP improvement to 46.7, using TensorRT improves it to 55.0, while upgrading the algorithm to TransFusion-L results in a lower L-mAP of 35.9.
These findings suggest that, within this budget constraint, implementing TensorRT for CenterPoint provides the best cost-performance.

When considering the case of deploying a single perception system, development costs become a more dominant factor compared to the deployment of multiple systems.
In this scenario, upgrading from an RTX4060Ti to an RTX3090 is the most cost-effective option, as it improves L-mAP to 46.7 at a relatively low additional cost.
In contrast, when deploying 100 perception systems, hardware costs become the primary concern.
Upgrading from an RTX4060Ti to an RTX3090 across all systems incurs a total cost of \$400k, leading to an L-mAP of 46.7.
However, switching from CenterPoint to TransFusion-L and utilizing TensorRT achieves a significantly higher L-mAP of 63.0 at a lower total cost of \$140k, making it the more cost-efficient choice.
These results demonstrate that the optimal strategy depends on the scale of deployment.
For small-scale deployments, upgrading the hardware provides the best balance between cost and performance, whereas for large-scale deployments, algorithmic optimization and acceleration using TensorRT yield the highest cost efficiency.
By quantitatively comparing different configurations using L-mAP, this study provides insights into the tradeoffs between hardware selection, algorithmic improvements, and deployment scale in industrial applications.

\section{Conclusion}
\label{sec:con}

In this paper, we propose latency-aware AP (L-AP) and planning-aware AP (P-AP) as new metrics, which consider the physical world such as the concept of time and physical constraints, offering a more comprehensive evaluation for real-time 3D object detection.
Using nuPlan dataset, we evaluate the effectiveness of these metrics for the whole autonomous driving system and demonstrate their utility in capturing planning-relevant aspects of 3D object detection.
Furthermore, we develop a state-of-the-art performance model for real-time 3D object detection through latency-aware hyperparameter optimization (L-HPO) based on our proposed metrics.
Our metrics also enable quantitative evaluation along a single axis, accounting for hardware differences and accelerators.
Additionally, we quantitatively optimize the number of frames for point clouds, as well as hardware and model selection, using our metrics.

We hope that this paper contributes to advancements in both algorithmic development within the research community and performance optimization for real-world applications in the industry.
As future work, we aim to further develop state-of-the-art models for real-time 3D object detection using the proposed metrics.
Additionally, we plan to refine these metrics for 3D object tracking, motion prediction, and motion planning to enhance real-time performance across a broader range of tasks.


{
    \small
    \bibliographystyle{ieeenat_fullname}
    \bibliography{main}

\begin{thebibliography}{53}
\providecommand{\natexlab}[1]{#1}
\providecommand{\url}[1]{\texttt{#1}}
\expandafter\ifx\csname urlstyle\endcsname\relax
  \providecommand{\doi}[1]{doi: #1}\else
  \providecommand{\doi}{doi: \begingroup \urlstyle{rm}\Url}\fi

\bibitem[NVI()]{NVIDIAAIIOT}
Nvidia-ai-iot.
\newblock \url{https://github.com/NVIDIA-AI-IOT/Lidar_AI_Solution}.

\bibitem[Ten()]{TensorRT}
Tensorrt.
\newblock \url{https://github.com/NVIDIA/TensorRT}.

\bibitem[Ali et~al.(2018)Ali, Abdelkarim, Zahran, Zidan, and
  Sallab]{Ali2018YOLO3DER}
Waleed Ali, Sherif Abdelkarim, Mohamed Zahran, Mahmoud Zidan, and Ahmad~El
  Sallab.
\newblock Yolo3d: End-to-end real-time 3d oriented object bounding box
  detection from lidar point cloud.
\newblock In \emph{ECCV Workshops}, 2018.

\bibitem[Bai et~al.(2022)Bai, Hu, Zhu, Huang, Chen, Fu, and
  Tai]{Bai2022TransFusionRL}
Xuyang Bai, Zeyu Hu, Xinge Zhu, Qingqiu Huang, Yilun Chen, Hongbo Fu, and
  Chiew-Lan Tai.
\newblock Transfusion: Robust lidar-camera fusion for 3d object detection with
  transformers.
\newblock \emph{2022 IEEE/CVF Conference on Computer Vision and Pattern
  Recognition (CVPR)}, pages 1080--1089, 2022.

\bibitem[Beemelmanns et~al.(2024)Beemelmanns, Zhang, Geller, and
  Eckstein]{Till2024}
Till Beemelmanns, Quan Zhang, Christian Geller, and Lutz Eckstein.
\newblock Multicorrupt: A multi-modal robustness dataset and benchmark of
  lidar-camera fusion for 3d object detection.
\newblock In \emph{2024 IEEE Intelligent Vehicles Symposium (IV)}, pages
  3255--3261, 2024.

\bibitem[Brazil et~al.(2023)Brazil, Kumar, Straub, Ravi, Johnson, and
  Gkioxari]{brazil2023omni3d}
Garrick Brazil, Abhinav Kumar, Julian Straub, Nikhila Ravi, Justin Johnson, and
  Georgia Gkioxari.
\newblock {Omni3D}: A large benchmark and model for {3D} object detection in
  the wild.
\newblock In \emph{CVPR}, Vancouver, Canada, 2023. IEEE.

\bibitem[Caesar et~al.(2020)Caesar, Bankiti, Lang, Vora, Liong, Xu, Krishnan,
  Pan, Baldan, and Beijbom]{nuscenes}
Holger Caesar, Varun Bankiti, Alex~H. Lang, Sourabh Vora, Venice~Erin Liong,
  Qiang Xu, Anush Krishnan, Yu Pan, Giancarlo Baldan, and Oscar Beijbom.
\newblock nuscenes: A multimodal dataset for autonomous driving.
\newblock In \emph{CVPR}, 2020.

\bibitem[Ceccarelli and Montecchi(2023)]{Ceccarelli2023}
Andrea Ceccarelli and Leonardo Montecchi.
\newblock Evaluating object (mis)detection from a safety and reliability
  perspective: Discussion and measures.
\newblock \emph{IEEE Access}, 11:\penalty0 44952--44963, 2023.

\bibitem[Chen et~al.(2020)Chen, Liu, Feng, Vallespi-Gonzalez, and
  Wellington]{Chen20203DPC}
Siheng Chen, Baoan Liu, Chen Feng, Carlos Vallespi-Gonzalez, and Carl~K.
  Wellington.
\newblock 3d point cloud processing and learning for autonomous driving:
  Impacting map creation, localization, and perception.
\newblock \emph{IEEE Signal Processing Magazine}, 38:\penalty0 68--86, 2020.

\bibitem[Chen et~al.(2016)Chen, Ma, Wan, Li, and Xia]{Chen2016Multiview3O}
Xiaozhi Chen, Huimin Ma, Ji Wan, Bo Li, and Tian Xia.
\newblock Multi-view 3d object detection network for autonomous driving.
\newblock \emph{2017 IEEE Conference on Computer Vision and Pattern Recognition
  (CVPR)}, pages 6526--6534, 2016.

\bibitem[Chen et~al.(2022)Chen, Zhang, Wang, Wang, and Zhao]{Chen2022FUTR3DAU}
Xuanyao Chen, Tianyuan Zhang, Yue Wang, Yilun Wang, and Hang Zhao.
\newblock Futr3d: A unified sensor fusion framework for 3d detection.
\newblock \emph{2023 IEEE/CVF Conference on Computer Vision and Pattern
  Recognition Workshops (CVPRW)}, pages 172--181, 2022.

\bibitem[Dauner et~al.(2023)Dauner, Hallgarten, Geiger, and
  Chitta]{Dauner2023CORL}
Daniel Dauner, Marcel Hallgarten, Andreas Geiger, and Kashyap Chitta.
\newblock Parting with misconceptions about learning-based vehicle motion
  planning.
\newblock In \emph{Conference on Robot Learning (CoRL)}, 2023.

\bibitem[Drews et~al.(2022)Drews, Feng, Faion, Rosenbaum, Ulrich, and
  Gl{\"a}ser]{Drews2022DeepFusionAR}
Florian Drews, Di Feng, Florian Faion, Lars Rosenbaum, Michael Ulrich, and
  Claudius Gl{\"a}ser.
\newblock Deepfusion: A robust and modular 3d object detector for lidars,
  cameras and radars.
\newblock \emph{2022 IEEE/RSJ International Conference on Intelligent Robots
  and Systems (IROS)}, pages 560--567, 2022.

\bibitem[Huang et~al.(2021)Huang, Huang, Zhu, and Du]{Huang2021BEVDetHM}
Junjie Huang, Guan Huang, Zheng Zhu, and Dalong Du.
\newblock Bevdet: High-performance multi-camera 3d object detection in
  bird-eye-view.
\newblock \emph{ArXiv}, abs/2112.11790, 2021.

\bibitem[Huang et~al.(2024)Huang, Ye, Liang, Shan, and Du]{Huang2024DL}
Junjie Huang, Yun Ye, Zhujin Liang, Yi Shan, and Dalong Du.
\newblock Detecting as labeling: Rethinking lidar-camera fusion in 3d object
  detection.
\newblock In \emph{Computer Vision ^^e2^^80^^93 ECCV 2024: 18th European
  Conference, Milan, Italy, September 29 ^^e2^^80^^93 October 4, 2024,
  Proceedings, Part XXII}, page 439^^e2^^80^^93455, Berlin, Heidelberg, 2024.
  Springer-Verlag.

\bibitem[Huang et~al.(2022)Huang, Shi, Li, Li, Huang, and
  Li]{Huang2022MultimodalSF}
Keli Huang, Botian Shi, Xiang Li, Xin Li, Siyuan Huang, and Yikang Li.
\newblock Multi-modal sensor fusion for auto driving perception: A survey.
\newblock \emph{ArXiv}, abs/2202.02703, 2022.

\bibitem[Hung et~al.(2024)Hung, Casser, Kretzschmar, Hwang, and
  Anguelov]{Hung2024}
Wei-Chih Hung, Vincent Casser, Henrik Kretzschmar, Jyh-Jing Hwang, and Dragomir
  Anguelov.
\newblock Let-3d-ap: Longitudinal error tolerant 3d average precision for
  camera-only 3d detection.
\newblock In \emph{2024 IEEE International Conference on Robotics and
  Automation (ICRA)}, pages 8272--8279, 2024.

\bibitem[Ivanovic and Pavone(2021)]{Ivanovic2021InjectingPI}
B. Ivanovic and Marco Pavone.
\newblock Injecting planning-awareness into prediction and detection
  evaluation.
\newblock \emph{2022 IEEE Intelligent Vehicles Symposium (IV)}, pages 821--828,
  2021.

\bibitem[Jiang et~al.(2024)Jiang, Li, Liu, Wang, Jia, Wang, Han, and
  Zhang]{JiangLLWJWHZ24}
Xiaohui Jiang, Shuailin Li, Yingfei Liu, Shihao Wang, Fan Jia, Tiancai Wang,
  Lijin Han, and Xiangyu Zhang.
\newblock Far3d: Expanding the horizon for surround-view 3d object detection.
\newblock In \emph{Thirty-Eighth {AAAI} Conference on Artificial Intelligence,
  {AAAI} 2024}, pages 2561--2569. {AAAI} Press, 2024.

\bibitem[Karnchanachari et~al.(2024)Karnchanachari, Geromichalos, Tan, Li,
  Eriksen, Yaghoubi, Mehdipour, Bernasconi, Fong, Guo, and
  Caesar]{Karnchanachari2024TowardsLP}
Napat Karnchanachari, Dimitris Geromichalos, Kok~Seang Tan, Nanxiang Li,
  Christopher Eriksen, Shakiba Yaghoubi, Noushin Mehdipour, Gianmarco
  Bernasconi, Whye~Kit Fong, Yiluan Guo, and Holger Caesar.
\newblock Towards learning-based planning: The nuplan benchmark for real-world
  autonomous driving.
\newblock \emph{2024 IEEE International Conference on Robotics and Automation
  (ICRA)}, pages 629--636, 2024.

\bibitem[Lang et~al.(2018)Lang, Vora, Caesar, Zhou, Yang, and
  Beijbom]{Lang2018PointPillarsFE}
Alex~H. Lang, Sourabh Vora, Holger Caesar, Lubing Zhou, Jiong Yang, and Oscar
  Beijbom.
\newblock Pointpillars: Fast encoders for object detection from point clouds.
\newblock \emph{2019 IEEE/CVF Conference on Computer Vision and Pattern
  Recognition (CVPR)}, pages 12689--12697, 2018.

\bibitem[Li et~al.(2022{\natexlab{a}})Li, Sima, Dai, Wang, Lu, Wang, Xie, Li,
  Deng, Tian, Zhu, Chen, Li, Gao, Geng, Zeng, Li, Yang, Jia, Yu, Qiao, Lin,
  Liu, Yan, Shi, and Luo]{Li2022DelvingIT}
Hongyang Li, Chonghao Sima, Jifeng Dai, Wenhai Wang, Lewei Lu, Huijie Wang,
  Enze Xie, Zhiqi Li, Hanming Deng, Haonan Tian, Xizhou Zhu, Li Chen, Tianyu
  Li, Yulu Gao, Xiangwei Geng, Jianqiang Zeng, Yang Li, Jiazhi Yang, Xiaosong
  Jia, Bo Yu, Y. Qiao, Dahua Lin, Siqian Liu, Junchi Yan, Jianping Shi, and
  Ping Luo.
\newblock Delving into the devils of bird’s-eye-view perception: A review,
  evaluation and recipe.
\newblock \emph{IEEE Transactions on Pattern Analysis and Machine
  Intelligence}, 46:\penalty0 2151--2170, 2022{\natexlab{a}}.

\bibitem[Li et~al.(2022{\natexlab{b}})Li, Ge, Yu, Yang, Wang, Shi, Sun, and
  Li]{Li2022BEVDepthAO}
Yinhao Li, Zheng Ge, Guanyi Yu, Jinrong Yang, Zengran Wang, Yukang Shi,
  Jian‐Yuan Sun, and Zeming Li.
\newblock Bevdepth: Acquisition of reliable depth for multi-view 3d object
  detection.
\newblock \emph{ArXiv}, abs/2206.10092, 2022{\natexlab{b}}.

\bibitem[Li et~al.(2022{\natexlab{c}})Li, Yu, Meng, Caine, Ngiam, Peng, Shen,
  Lu, Zhou, Le, Yuille, and Tan]{Yingwei2022}
Yingwei Li, Adams~Wei Yu, Tianjian Meng, Ben Caine, Jiquan Ngiam, Daiyi Peng,
  Junyang Shen, Yifeng Lu, Denny Zhou, Quoc~V. Le, Alan Yuille, and Mingxing
  Tan.
\newblock Deepfusion: Lidar-camera deep fusion for multi-modal 3d object
  detection.
\newblock In \emph{2022 IEEE/CVF Conference on Computer Vision and Pattern
  Recognition (CVPR)}, pages 17161--17170, 2022{\natexlab{c}}.

\bibitem[Li et~al.(2024)Li, Fan, Liu, Huang, Chen, Wang, and
  Zhang]{li2024fully}
Yingyan Li, Lue Fan, Yang Liu, Zehao Huang, Yuntao Chen, Naiyan Wang, and
  Zhaoxiang Zhang.
\newblock Fully sparse fusion for 3d object detection.
\newblock \emph{IEEE Transactions on Pattern Analysis and Machine
  Intelligence}, 2024.

\bibitem[Li et~al.(2022{\natexlab{d}})Li, Wang, Li, Xie, Sima, Lu, Qiao, and
  Dai]{li2022bevformer}
Zhiqi Li, Wenhai Wang, Hongyang Li, Enze Xie, Chonghao Sima, Tong Lu, Yu Qiao,
  and Jifeng Dai.
\newblock Bevformer: Learning bird’s-eye-view representation from multi-camera
  images via spatiotemporal transformers.
\newblock pages 1--18, 2022{\natexlab{d}}.

\bibitem[Liang et~al.(2022)Liang, Xie, Yu, Xia, Lin, Wang, Tang, Wang, and
  Tang]{Liang2022BEVFusionAS}
Tingting Liang, Hongwei Xie, Kaicheng Yu, Zhongyu Xia, Zhiwei Lin, Yongtao
  Wang, Tao Tang, Bing Wang, and Zhi Tang.
\newblock Bevfusion: A simple and robust lidar-camera fusion framework.
\newblock \emph{ArXiv}, abs/2205.13790, 2022.

\bibitem[Liu et~al.(2022)Liu, Leng, Sun, Cheng, Qi, Zhou, Tan, and
  Anguelov]{Liu2022}
Chenxi Liu, Zhaoqi Leng, Pei Sun, Shuyang Cheng, Charles~R. Qi, Yin Zhou,
  Mingxing Tan, and Dragomir Anguelov.
\newblock Lidarnas: Unifying and searching neural architectures for 3d point
  clouds.
\newblock In \emph{Computer Vision ^^e2^^80^^93 ECCV 2022: 17th European
  Conference, Tel Aviv, Israel, October 23^^e2^^80^^9327, 2022, Proceedings,
  Part XXI}, page 158^^e2^^80^^93175, Berlin, Heidelberg, 2022.
  Springer-Verlag.

\bibitem[Liu et~al.(2023)Liu, Tang, Amini, Yang, Mao, Rus, and
  Han]{liu2022bevfusion}
Zhijian Liu, Haotian Tang, Alexander Amini, Xingyu Yang, Huizi Mao, Daniela
  Rus, and Song Han.
\newblock Bevfusion: Multi-task multi-sensor fusion with unified bird's-eye
  view representation.
\newblock In \emph{IEEE International Conference on Robotics and Automation
  (ICRA)}, 2023.

\bibitem[Mao et~al.(2023)Mao, Shi, Wang, and
  Li]{mao20233dobjectdetectionautonomous}
Jiageng Mao, Shaoshuai Shi, Xiaogang Wang, and Hongsheng Li.
\newblock 3d object detection for autonomous driving: A comprehensive survey,
  2023.

\bibitem[Oboril et~al.(2022)Oboril, B{\"u}rkle, Sussmann, Bitton, and
  Fabris]{Oboril2022MTBFMF}
Fabian Oboril, Cornelius B{\"u}rkle, Alon Sussmann, Simcha Bitton, and Simone
  Fabris.
\newblock Mtbf model for avs - from perception errors to vehicle-level
  failures.
\newblock \emph{2022 IEEE Intelligent Vehicles Symposium (IV)}, pages
  1591--1598, 2022.

\bibitem[Philion and Fidler(2020)]{Philion2020}
Jonah Philion and Sanja Fidler.
\newblock Lift, splat, shoot: Encoding images from arbitrary camera rigs by
  implicitly unprojecting to 3d.
\newblock In \emph{Computer Vision ^^e2^^80^^93 ECCV 2020: 16th European
  Conference, Glasgow, UK, August 23^^e2^^80^^9328, 2020, Proceedings, Part
  XIV}, page 194^^e2^^80^^93210, Berlin, Heidelberg, 2020. Springer-Verlag.

\bibitem[Qi et~al.(2016)Qi, Su, Mo, and Guibas]{Qi2016PointNetDL}
C. Qi, Hao Su, Kaichun Mo, and Leonidas~J. Guibas.
\newblock Pointnet: Deep learning on point sets for 3d classification and
  segmentation.
\newblock \emph{2017 IEEE Conference on Computer Vision and Pattern Recognition
  (CVPR)}, pages 77--85, 2016.

\bibitem[Qi et~al.(2017)Qi, Yi, Su, and Guibas]{Qi2017}
Charles~R. Qi, Li Yi, Hao Su, and Leonidas~J. Guibas.
\newblock Pointnet++: deep hierarchical feature learning on point sets in a
  metric space.
\newblock In \emph{Proceedings of the 31st International Conference on Neural
  Information Processing Systems}, page 5105^^e2^^80^^935114, Red Hook, NY,
  USA, 2017. Curran Associates Inc.

\bibitem[Qi et~al.(2018)Qi, Liu, Wu, Su, and Guibas]{Qi2018}
Charles~R. Qi, Wei Liu, Chenxia Wu, Hao Su, and Leonidas~J. Guibas.
\newblock Frustum pointnets for 3d object detection from rgb-d data.
\newblock In \emph{2018 IEEE/CVF Conference on Computer Vision and Pattern
  Recognition}, pages 918--927, 2018.

\bibitem[Shi et~al.(2018)Shi, Wang, and Li]{Shi2018PointRCNN3O}
Shaoshuai Shi, Xiaogang Wang, and Hongsheng Li.
\newblock Pointrcnn: 3d object proposal generation and detection from point
  cloud.
\newblock \emph{2019 IEEE/CVF Conference on Computer Vision and Pattern
  Recognition (CVPR)}, pages 770--779, 2018.

\bibitem[Shi et~al.(2019)Shi, Guo, Jiang, Wang, Shi, Wang, and
  Li]{Shi2019PVRCNNPF}
Shaoshuai Shi, Chaoxu Guo, Li Jiang, Zhe Wang, Jianping Shi, Xiaogang Wang, and
  Hongsheng Li.
\newblock Pv-rcnn: Point-voxel feature set abstraction for 3d object detection.
\newblock \emph{2020 IEEE/CVF Conference on Computer Vision and Pattern
  Recognition (CVPR)}, pages 10526--10535, 2019.

\bibitem[Sun et~al.(2020)Sun, Kretzschmar, Dotiwalla, Chouard, Patnaik, Tsui,
  Guo, Zhou, Chai, Caine, Vasudevan, Han, Ngiam, Zhao, Timofeev, Ettinger,
  Krivokon, Gao, Joshi, Zhang, Shlens, Chen, and Anguelov]{Waymo2020}
Pei Sun, Henrik Kretzschmar, Xerxes Dotiwalla, Aurelien Chouard, Vijaysai
  Patnaik, Paul Tsui, James Guo, Yin Zhou, Yuning Chai, Benjamin Caine, Vijay
  Vasudevan, Wei Han, Jiquan Ngiam, Hang Zhao, Aleksei Timofeev, Scott
  Ettinger, Maxim Krivokon, Amy Gao, Aditya Joshi, Yu Zhang, Jonathon Shlens,
  Zhifeng Chen, and Dragomir Anguelov.
\newblock Scalability in perception for autonomous driving: Waymo open dataset.
\newblock In \emph{2020 {IEEE/CVF} Conference on Computer Vision and Pattern
  Recognition, {CVPR} 2020, Seattle, WA, USA, June 13-19, 2020}, pages
  2443--2451. Computer Vision Foundation / {IEEE}, 2020.

\bibitem[Tian et~al.(2019)Tian, Shen, Chen, and He]{Tian2019FCOSFC}
Zhi Tian, Chunhua Shen, Hao Chen, and Tong He.
\newblock Fcos: Fully convolutional one-stage object detection.
\newblock \emph{2019 IEEE/CVF International Conference on Computer Vision
  (ICCV)}, pages 9626--9635, 2019.

\bibitem[Vora et~al.(2019)Vora, Lang, Helou, and
  Beijbom]{Vora2019PointPaintingSF}
Sourabh Vora, Alex~H. Lang, Bassam Helou, and Oscar Beijbom.
\newblock Pointpainting: Sequential fusion for 3d object detection.
\newblock \emph{2020 IEEE/CVF Conference on Computer Vision and Pattern
  Recognition (CVPR)}, pages 4603--4611, 2019.

\bibitem[Wang et~al.(2021{\natexlab{a}})Wang, Ma, Zhu, and
  Yang]{wang2021pointaugmenting}
Chunwei Wang, Chao Ma, Ming Zhu, and Xiaokang Yang.
\newblock Pointaugmenting: Cross-modal augmentation for 3d object detection.
\newblock In \emph{Proceedings of the IEEE/CVF Conference on Computer Vision
  and Pattern Recognition}, pages 11794--11803, 2021{\natexlab{a}}.

\bibitem[Wang et~al.(2024)Wang, Meng, Liu, Yan, Wang, Cheng, and
  Hou]{WangMLYWCH24}
Jiabao Wang, Qiang Meng, Guochao Liu, Liujiang Yan, Ke Wang, Ming-Ming Cheng,
  and Qibin Hou.
\newblock Towards stable 3d object detection.
\newblock In \emph{Computer Vision ^^e2^^80^^93 ECCV 2024: 18th European
  Conference, Milan, Italy, September 29^^e2^^80^^93October 4, 2024,
  Proceedings, Part L}, page 197^^e2^^80^^93213, Berlin, Heidelberg, 2024.
  Springer-Verlag.

\bibitem[Wang et~al.(2023)Wang, Liu, Wang, Li, and Zhang]{Wang2023ExploringOT}
Shihao Wang, Yingfei Liu, Tiancai Wang, Ying Li, and Xiangyu Zhang.
\newblock Exploring object-centric temporal modeling for efficient multi-view
  3d object detection.
\newblock \emph{2023 IEEE/CVF International Conference on Computer Vision
  (ICCV)}, pages 3598--3608, 2023.

\bibitem[Wang and Kitani(2022)]{Wang2022CostAwareEA}
Xiaofang Wang and Kris~M. Kitani.
\newblock Cost-aware evaluation and model scaling for lidar-based 3d object
  detection.
\newblock \emph{2023 IEEE International Conference on Robotics and Automation
  (ICRA)}, pages 9260--9267, 2022.

\bibitem[Wang et~al.(2021{\natexlab{b}})Wang, Guizilini, Zhang, Wang, Zhao, ,
  and Solomon]{detr3d}
Yue Wang, Vitor Guizilini, Tianyuan Zhang, Yilun Wang, Hang Zhao, , and
  Justin~M. Solomon.
\newblock Detr3d: 3d object detection from multi-view images via 3d-to-2d
  queries.
\newblock In \emph{The Conference on Robot Learning ({CoRL})},
  2021{\natexlab{b}}.

\bibitem[Yan et~al.(2018)Yan, Mao, and Li]{Yan2018SECONDSE}
Yan Yan, Yuxing Mao, and Bo Li.
\newblock Second: Sparsely embedded convolutional detection.
\newblock \emph{Sensors (Basel, Switzerland)}, 18, 2018.

\bibitem[Yang et~al.(2018)Yang, Luo, and Urtasun]{Yang2018PIXORR3}
Binh Yang, Wenjie Luo, and Raquel Urtasun.
\newblock Pixor: Real-time 3d object detection from point clouds.
\newblock \emph{2018 IEEE/CVF Conference on Computer Vision and Pattern
  Recognition}, pages 7652--7660, 2018.

\bibitem[Yang et~al.(2023)Yang, Chen, Tian, Tao, Zhu, Zhang, Huang, Li, Qiao,
  Lu, Zhou, and Dai]{Yang2022BEVFormerVA}
Chenyu Yang, Yuntao Chen, Hao Tian, Chenxin Tao, Xizhou Zhu, Zhaoxiang Zhang,
  Gao Huang, Hongyang Li, Yu Qiao, Lewei Lu, Jie Zhou, and Jifeng Dai.
\newblock { BEVFormer v2: Adapting Modern Image Backbones to Bird's-Eye-View
  Recognition via Perspective Supervision }.
\newblock pages 17830--17839, 2023.

\bibitem[Yang et~al.(2020)Yang, Sun, Liu, and Jia]{Yang20203DSSDP3}
Zetong Yang, Yanan Sun, Shu Liu, and Jiaya Jia.
\newblock 3dssd: Point-based 3d single stage object detector.
\newblock \emph{2020 IEEE/CVF Conference on Computer Vision and Pattern
  Recognition (CVPR)}, pages 11037--11045, 2020.

\bibitem[Yang et~al.(2024)Yang, Yu, Choy, Wang, Anandkumar, and
  Alvarez]{Yang2024}
Zetong Yang, Zhiding Yu, Chris Choy, Renhao Wang, Anima Anandkumar, and Jose~M.
  Alvarez.
\newblock Improving distant 3d object detection using 2d box supervision.
\newblock In \emph{2024 IEEE/CVF Conference on Computer Vision and Pattern
  Recognition (CVPR)}, pages 14853--14863, 2024.

\bibitem[Yin et~al.(2021)Yin, Zhou, and Kr{\"a}henb{\"u}hl]{yin2021center}
Tianwei Yin, Xingyi Zhou, and Philipp Kr{\"a}henb{\"u}hl.
\newblock Center-based 3d object detection and tracking.
\newblock \emph{CVPR}, 2021.

\bibitem[Zhang et~al.(2022)Zhang, Zhu, Zheng, Huang, Huang, Zhou, and
  Lu]{Zhang2022BEVerseUP}
Yunpeng Zhang, Zheng~Hua Zhu, Wenzhao Zheng, Junjie Huang, Guan Huang, Jie
  Zhou, and Jiwen Lu.
\newblock Beverse: Unified perception and prediction in birds-eye-view for
  vision-centric autonomous driving.
\newblock \emph{ArXiv}, abs/2205.09743, 2022.

\bibitem[Zhou and Tuzel(2017)]{Zhou2017VoxelNetEL}
Yin Zhou and Oncel Tuzel.
\newblock Voxelnet: End-to-end learning for point cloud based 3d object
  detection.
\newblock \emph{2018 IEEE/CVF Conference on Computer Vision and Pattern
  Recognition}, pages 4490--4499, 2017.

\end{thebibliography}
}

\clearpage
\setcounter{page}{1}
\maketitlesupplementary
\section{Appendix}

\subsection{L-AP for Yaw Flipping}
\label{sec:appendix-1}

\begin{figure}[h]
    \begin{center}
        \includegraphics[width=0.8\linewidth]{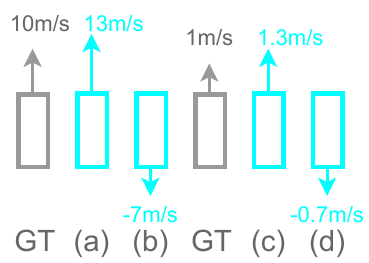}
        \caption{
          The case to evaluate for yaw flipping.
          (a) The velocity estimation is sufficiently accurate.
          (b) The direction is estimated in the opposite orientation.
          (c) The object is almost stationary.
          (d) The object is almost stationary but estimated in the opposite direction.
        }
        \label{flipping-figure}
    \end{center}
\end{figure}

\begin{table}[h]
    \centering
    \caption{
      Comparison of mAP and L-mAP at different inference time with $\pi$ yaw difference.
      The thresholds are (0.5, 1.0, 1.5, 2.0).
    }
    \begin{tabular}{|c|c|c|c|c|} \hline
        & (a)  & (b)  & (c)  & (d) \\ \hline
        mAP           & 1.0  & 1.0  & 1.0  & 1.0  \\ \hline
        L-mAP for 100ms  & 1.0  & 0.25 & 1.0  & 1.0  \\ \hline
        L-mAP for 200ms  & 0.75 & 0.0  & 1.0  & 1.0  \\ \hline
        L-mAP for 1000ms & 0.0  & 0.0  & 1.0  & 0.25 \\ \hline
    \end{tabular}
    \label{flipping-comparison}
\end{table}

By introducing latency into the evaluation, it becomes possible to assess objects whose yaw orientation is incorrectly estimated in the opposite direction.
In the cases of \cref{flipping-figure}, we compare the performance of L-mAP with different inference time as shown in \Cref{flipping-comparison}.
Comparing (a) and (c), the greater the relative velocity of the target, the more significantly latency impacts recognition performance degradation.
In (b), where the yaw is reversed by 180 degrees, the AP remains 1.0 when latency is not considered.
However, when latency is taken into account, the model properly reflects the performance degradation.
In (d), since the velocity is small, the performance degradation is less severe compared to (b).

\subsection{Annotation Frequency}
\label{sec:appendix-2}

\begin{figure*}[t]
  \begin{tabular}{cc}
    \begin{minipage}[t]{0.40\hsize}
      \centering
      \includegraphics[width=0.99\linewidth]{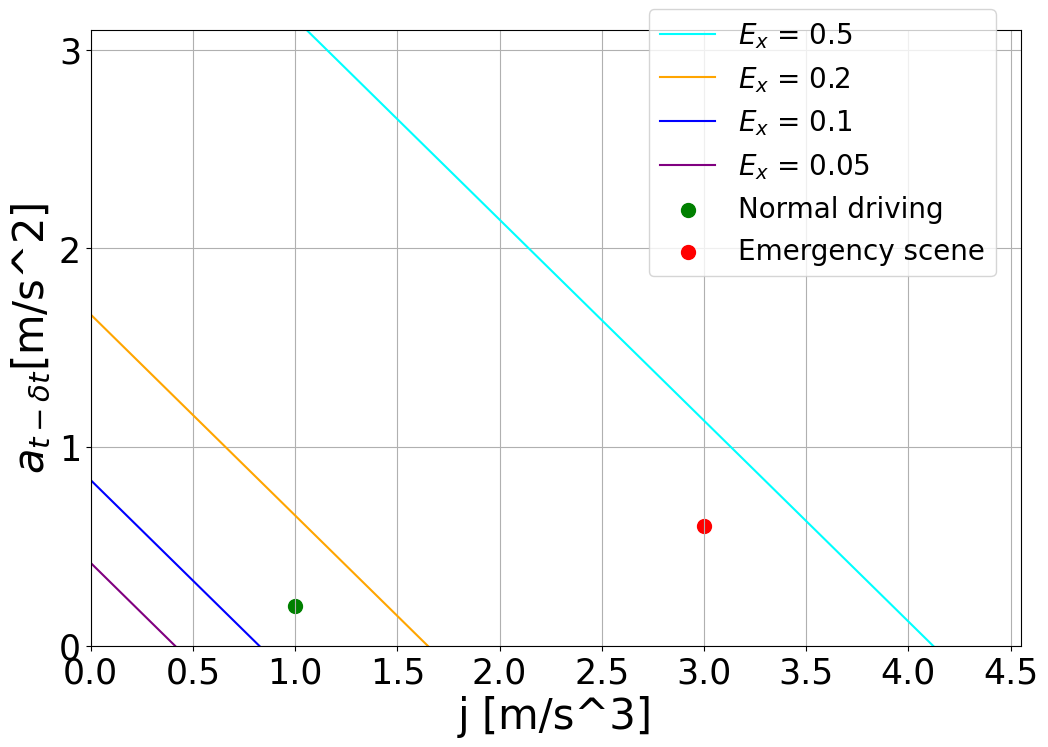}
      \hspace{1.6cm} (a) $\delta t = 1.0 $ (1 Hz annotation).
    \end{minipage} &
    \begin{minipage}[t]{0.40\hsize}
      \centering
      \includegraphics[width=0.99\linewidth]{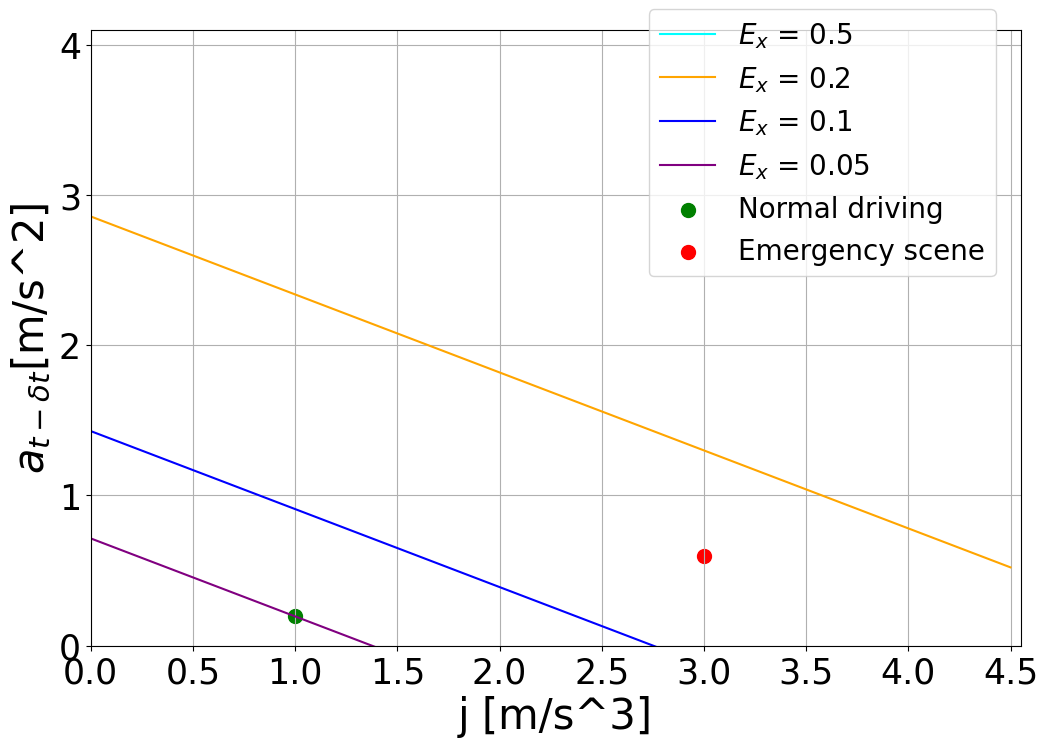}
      \hspace{1.6cm} (b) $\delta t = 0.5 $ (2 Hz annotation).
    \end{minipage} \\
    \begin{minipage}[t]{0.40\hsize}
      \centering
      \includegraphics[width=0.99\linewidth]{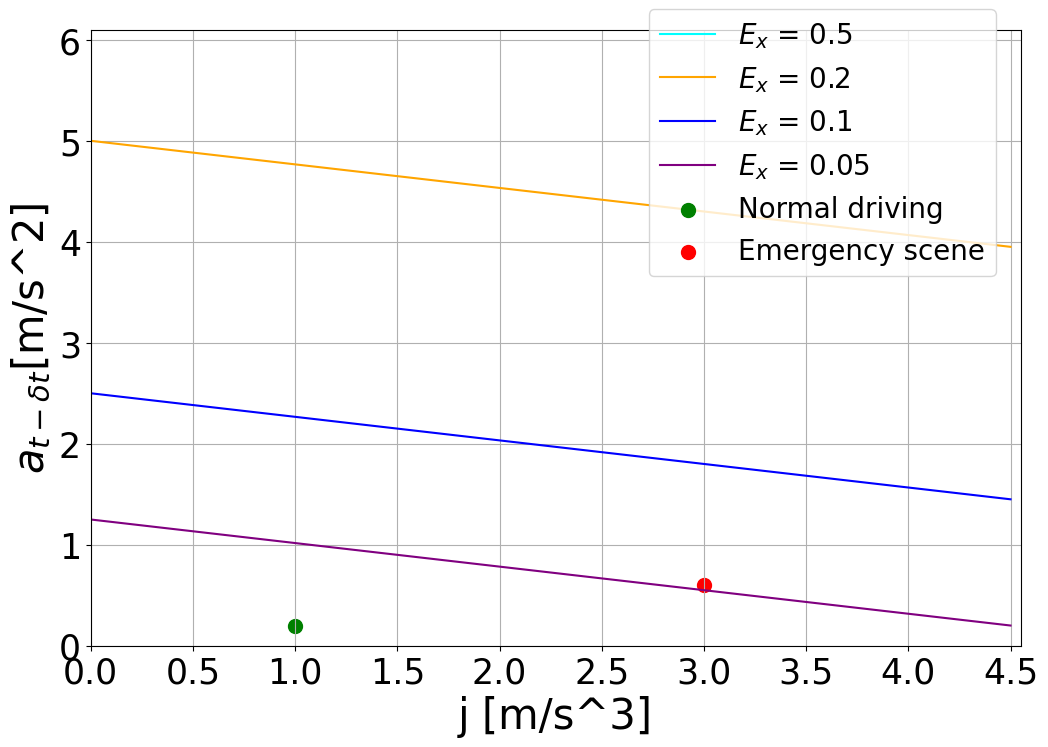}
      \hspace{1.6cm} (c) $\delta t = 0.2 $ (5 Hz annotation).
    \end{minipage} &
    \begin{minipage}[t]{0.40\hsize}
      \centering
      \includegraphics[width=0.99\linewidth]{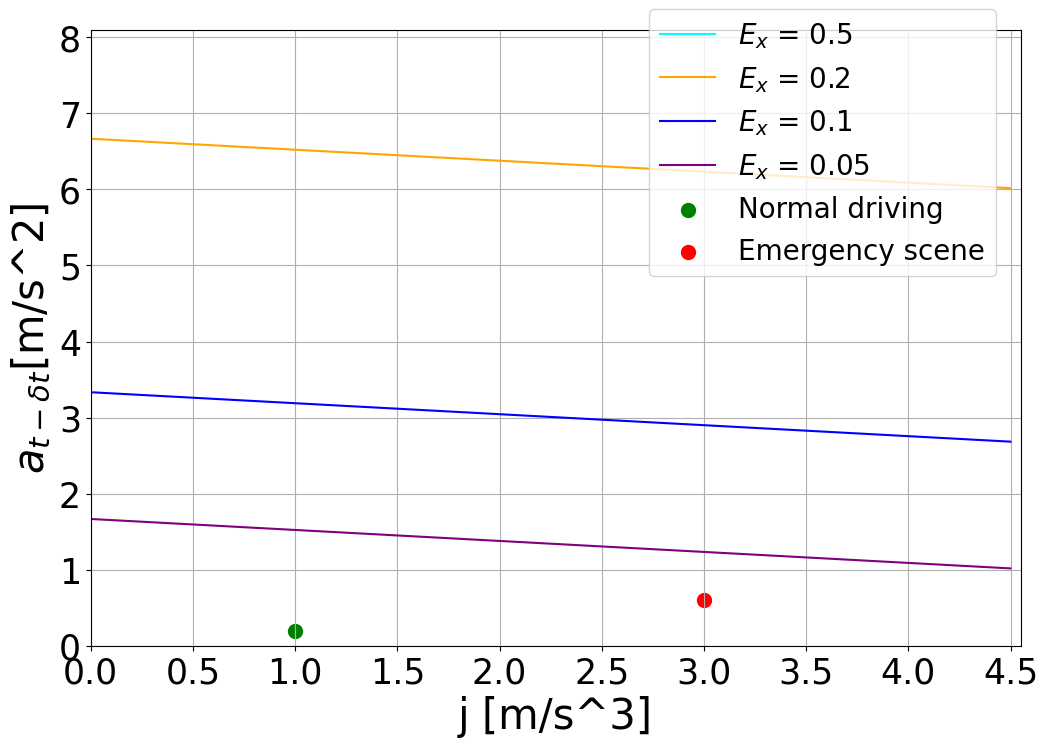}
      \hspace{1.6cm} (d) $\delta t = 0.1 $ (10 Hz annotation).
    \end{minipage}
  \end{tabular}
  \caption{
    The estimation error of $ x_{t + \Delta t}^{GT} $ for inference time $\Delta t$ = 0.2 s.
  }
  \label{x_DT02}
\end{figure*}

\begin{figure*}[th]
  \begin{tabular}{cc}
    \begin{minipage}[t]{0.40\hsize}
      \centering
      \includegraphics[width=0.9\linewidth]{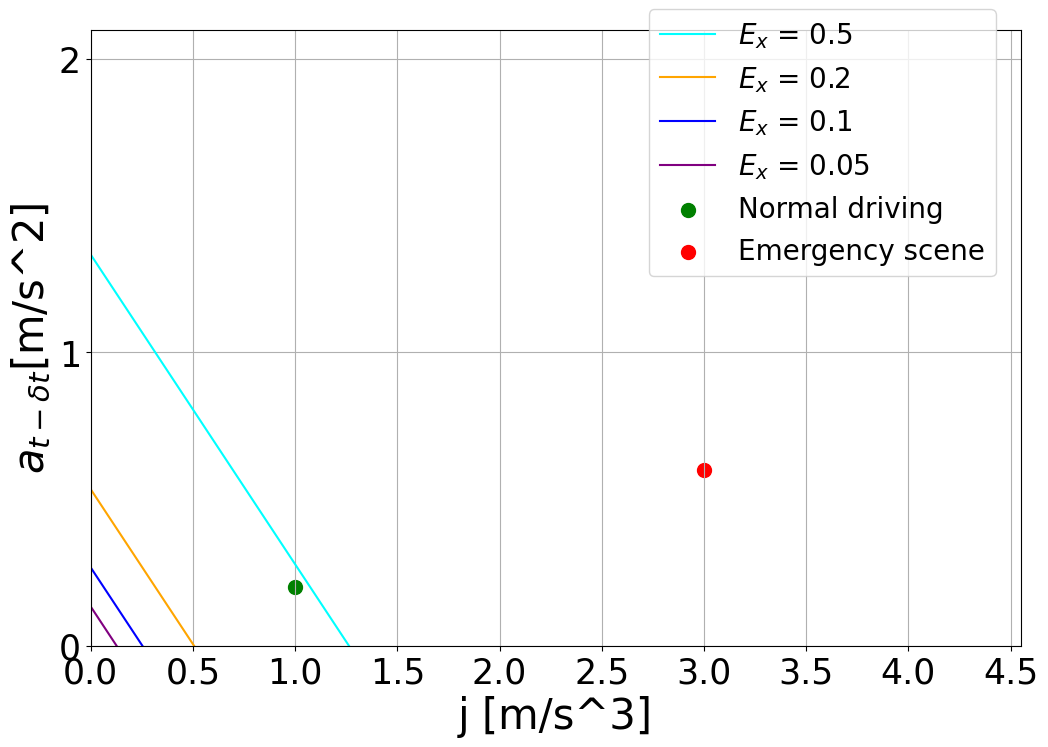}
      \hspace{1.6cm} (a) $\delta t = 1.0 $ (1 Hz annotation).
    \end{minipage} &
    \begin{minipage}[t]{0.40\hsize}
      \centering
      \includegraphics[width=0.9\linewidth]{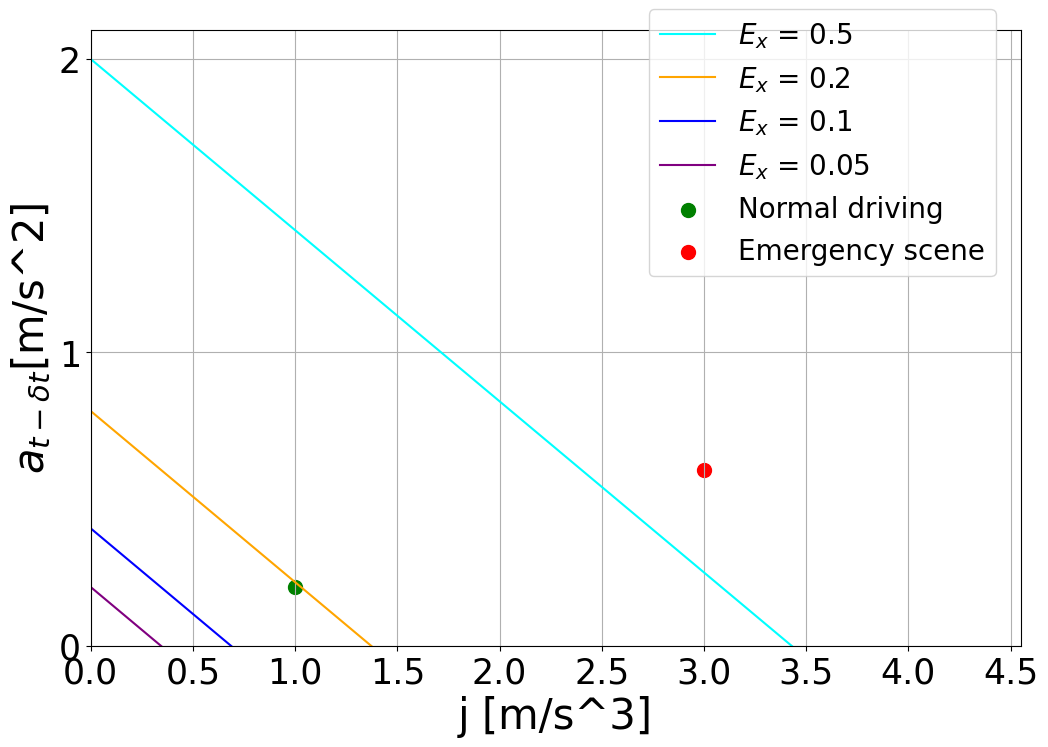}
      \hspace{1.6cm} (b) $\delta t = 0.5 $ (2 Hz annotation).
    \end{minipage} \\
    \begin{minipage}[t]{0.40\hsize}
      \centering
      \includegraphics[width=0.9\linewidth]{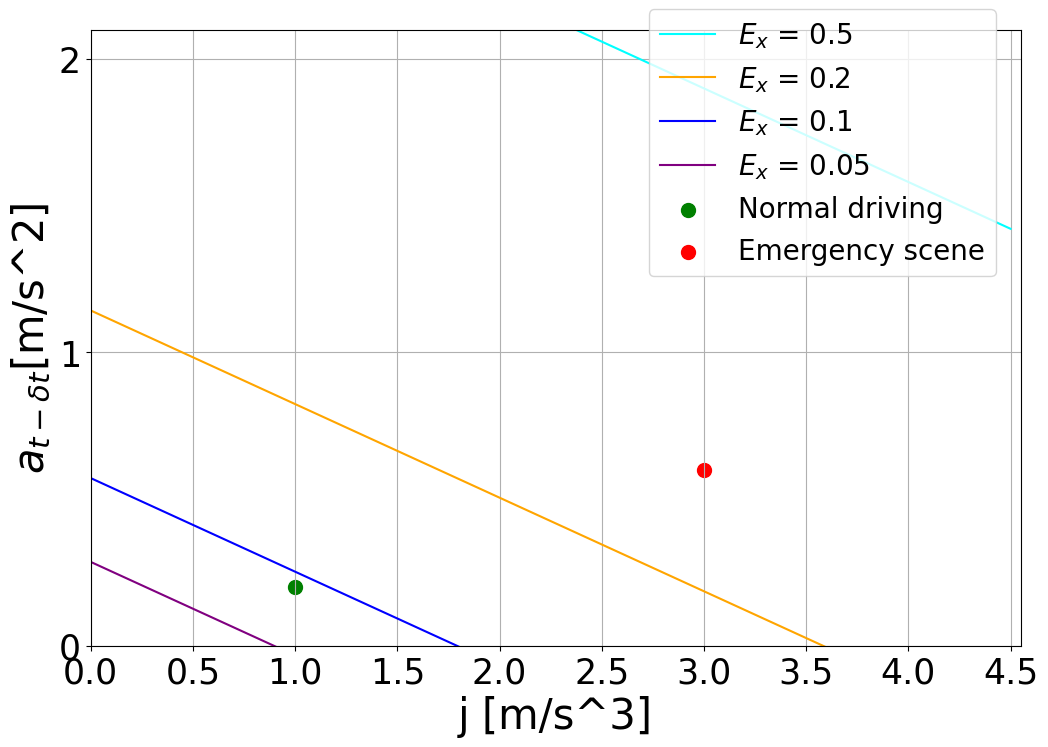}
      \hspace{1.6cm} (c) $\delta t = 0.2 $ (5 Hz annotation).
    \end{minipage} &
    \begin{minipage}[t]{0.40\hsize}
      \centering
      \includegraphics[width=0.9\linewidth]{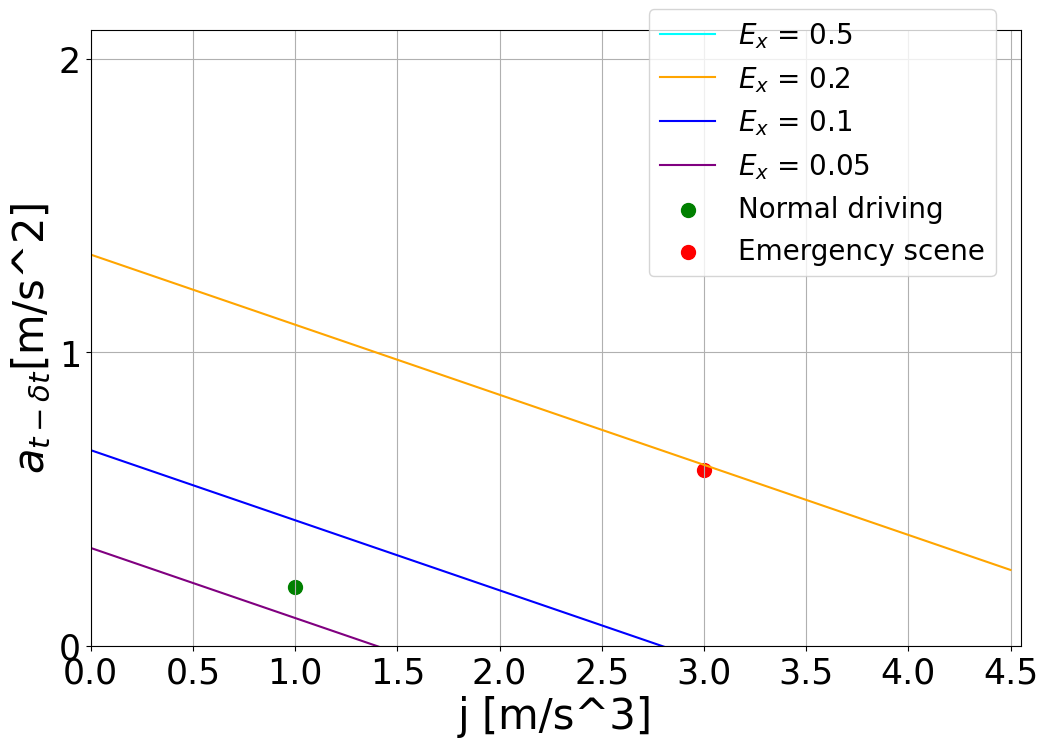}
      \hspace{1.6cm} (d) $\delta t = 0.1 $ (10 Hz annotation).
    \end{minipage}
  \end{tabular}
  \caption{
    The estimation error of $ x_{t + \Delta t}^{GT} $ for inference time $\Delta t$ = 0.5 s.
  }
  \label{x_DT05}
\end{figure*}

\begin{figure*}[th]
  \begin{tabular}{cc}
    \begin{minipage}[t]{0.40\hsize}
      \centering
      \includegraphics[width=0.9\linewidth]{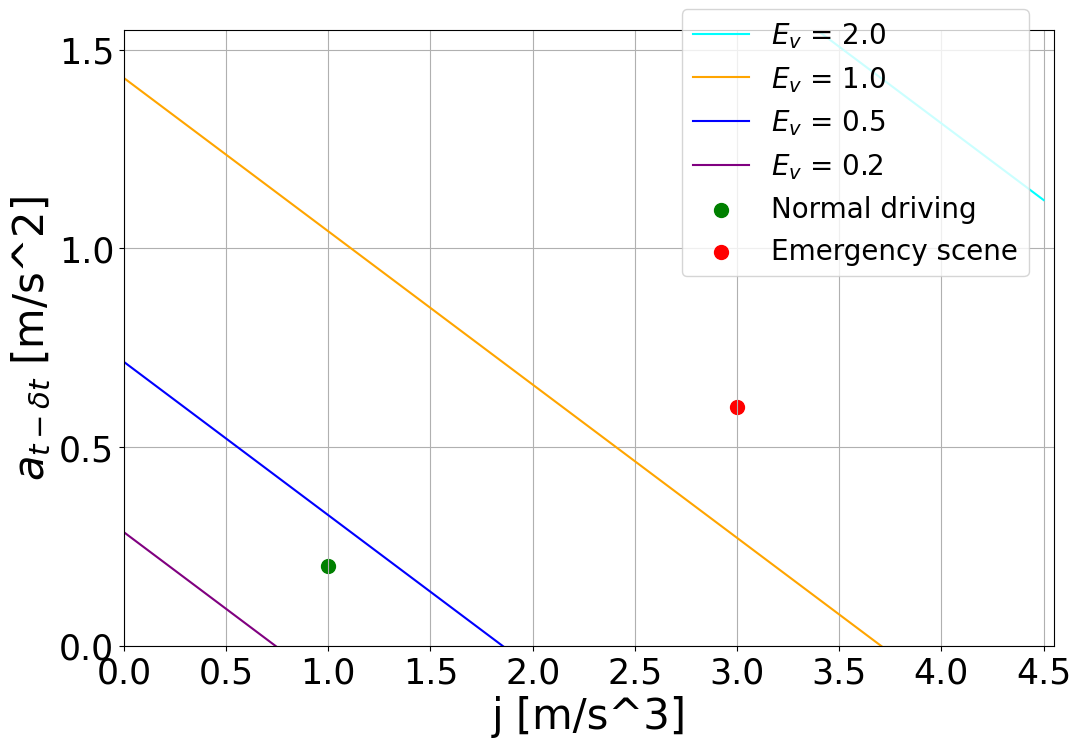}
      \hspace{1.6cm} (a) $\delta t = 1.0 $ (1 Hz annotation).
    \end{minipage} &
    \begin{minipage}[t]{0.40\hsize}
      \centering
      \includegraphics[width=0.9\linewidth]{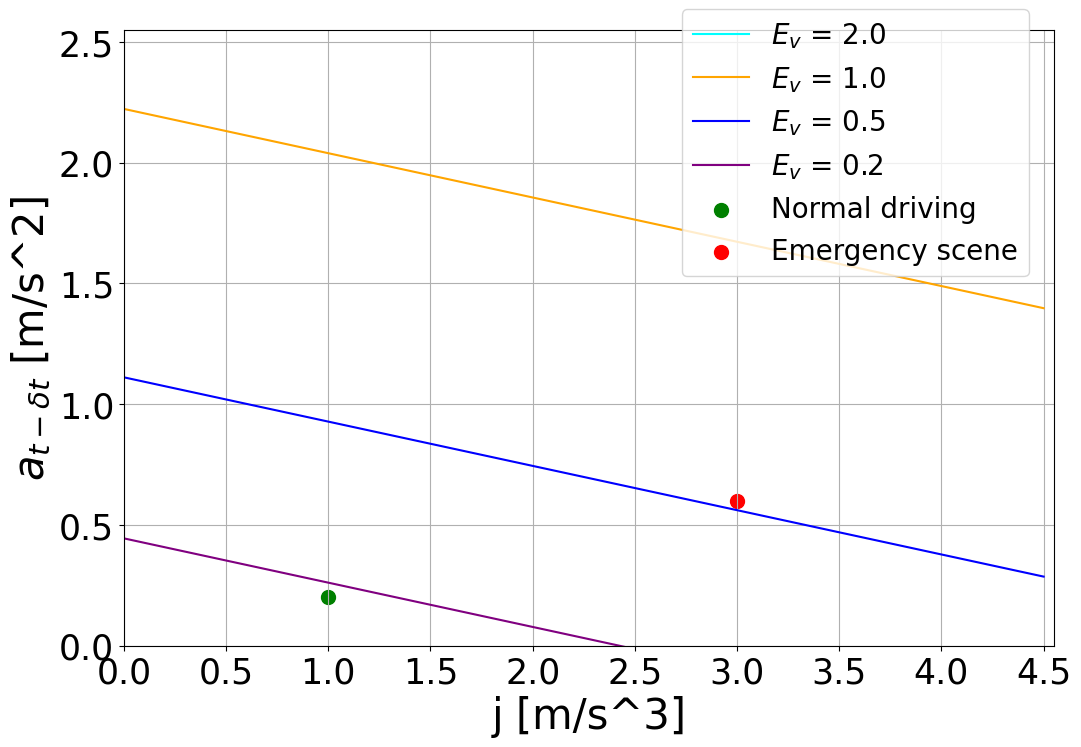}
      \hspace{1.6cm} (b) $\delta t = 0.5 $ (2 Hz annotation).
    \end{minipage} \\
    \begin{minipage}[t]{0.40\hsize}
      \centering
      \includegraphics[width=0.9\linewidth]{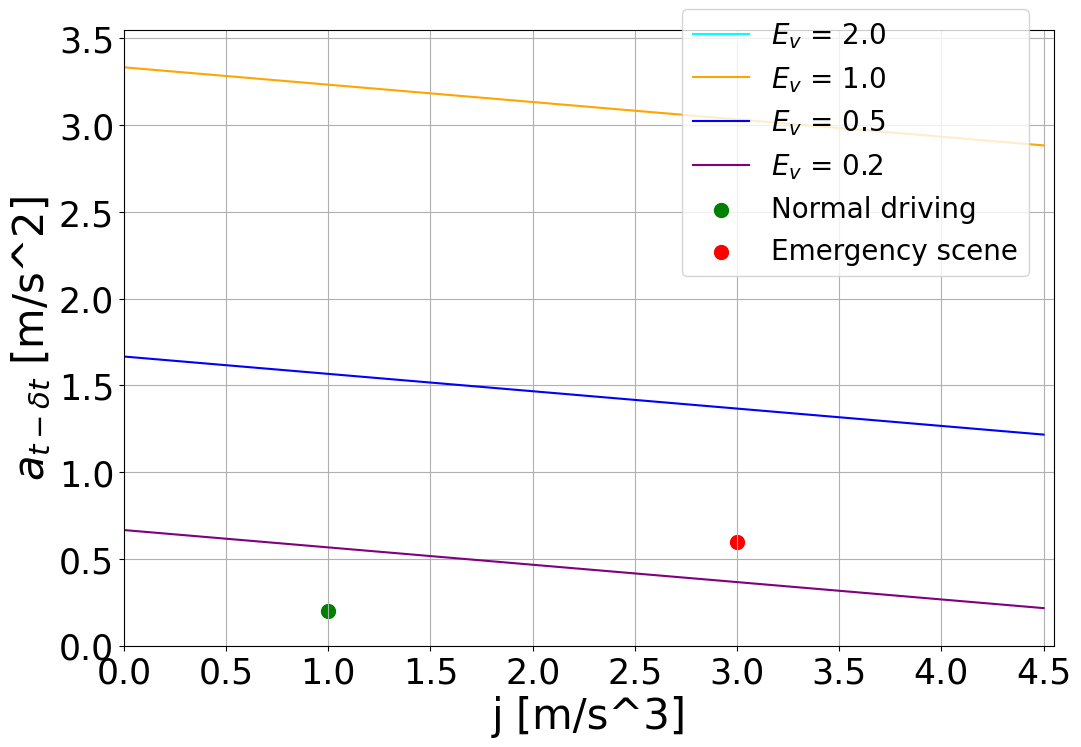}
      \hspace{1.6cm} (c) $\delta t = 0.2 $ (5 Hz annotation).
    \end{minipage} &
    \begin{minipage}[t]{0.40\hsize}
      \centering
      \includegraphics[width=0.9\linewidth]{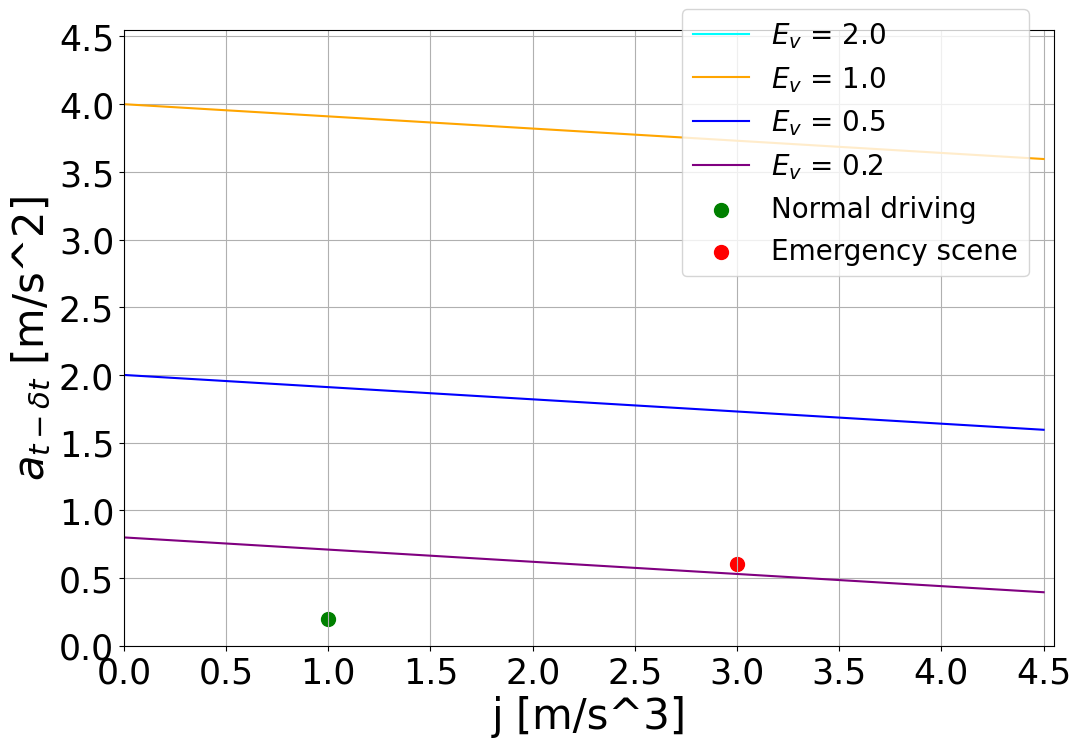}
      \hspace{1.6cm} (d) $\delta t = 0.1 $ (10 Hz annotation).
    \end{minipage}
  \end{tabular}
  \caption{
    The estimation error of $ v_{t + \Delta t}^{GT} $ for inference time $\Delta t$ = 0.2 s.
  }
  \label{v_DT02}
\end{figure*}

\begin{figure*}[th]
  \begin{tabular}{cc}
    \begin{minipage}[t]{0.40\hsize}
      \centering
      \includegraphics[width=0.9\linewidth]{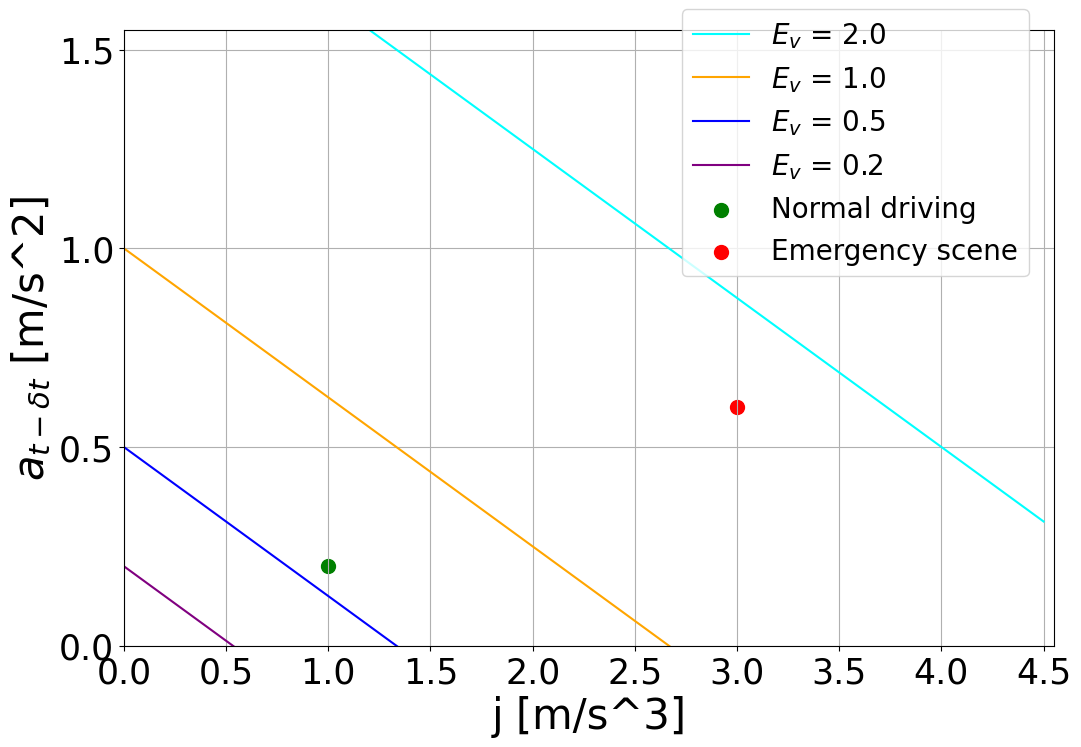}
      \hspace{1.6cm} (a) $\delta t = 1.0 $ (1 Hz annotation).
    \end{minipage} &
    \begin{minipage}[t]{0.40\hsize}
      \centering
      \includegraphics[width=0.9\linewidth]{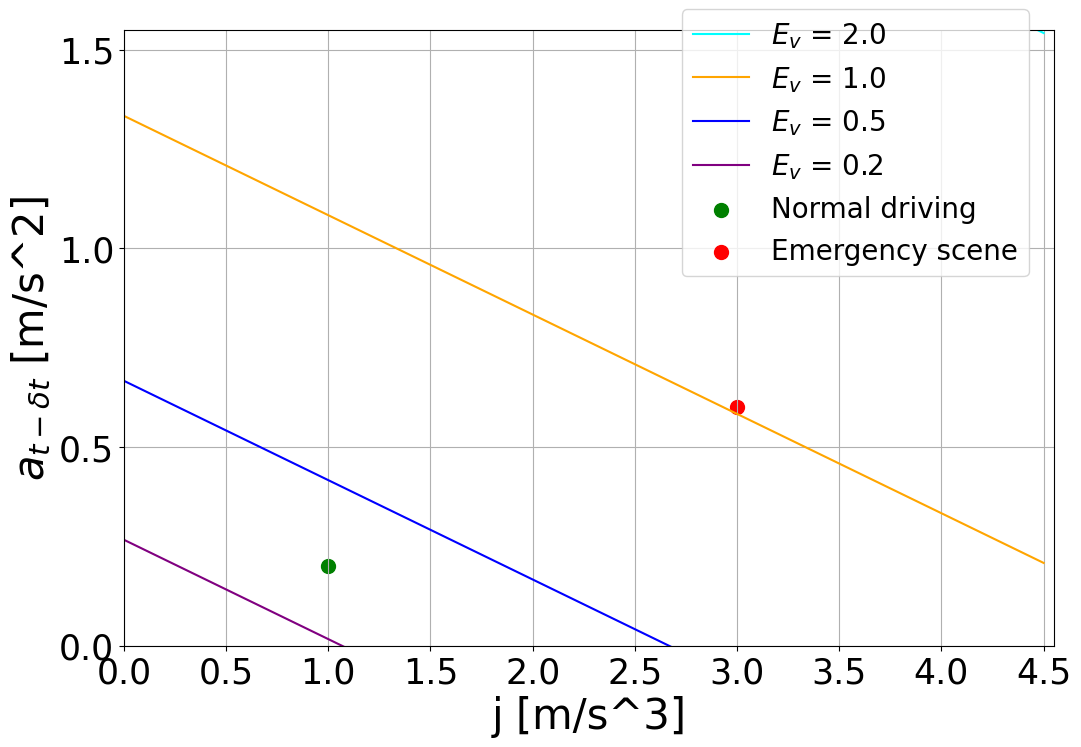}
      \hspace{1.6cm} (b) $\delta t = 0.5 $ (2 Hz annotation).
    \end{minipage} \\
    \begin{minipage}[t]{0.40\hsize}
      \centering
      \includegraphics[width=0.9\linewidth]{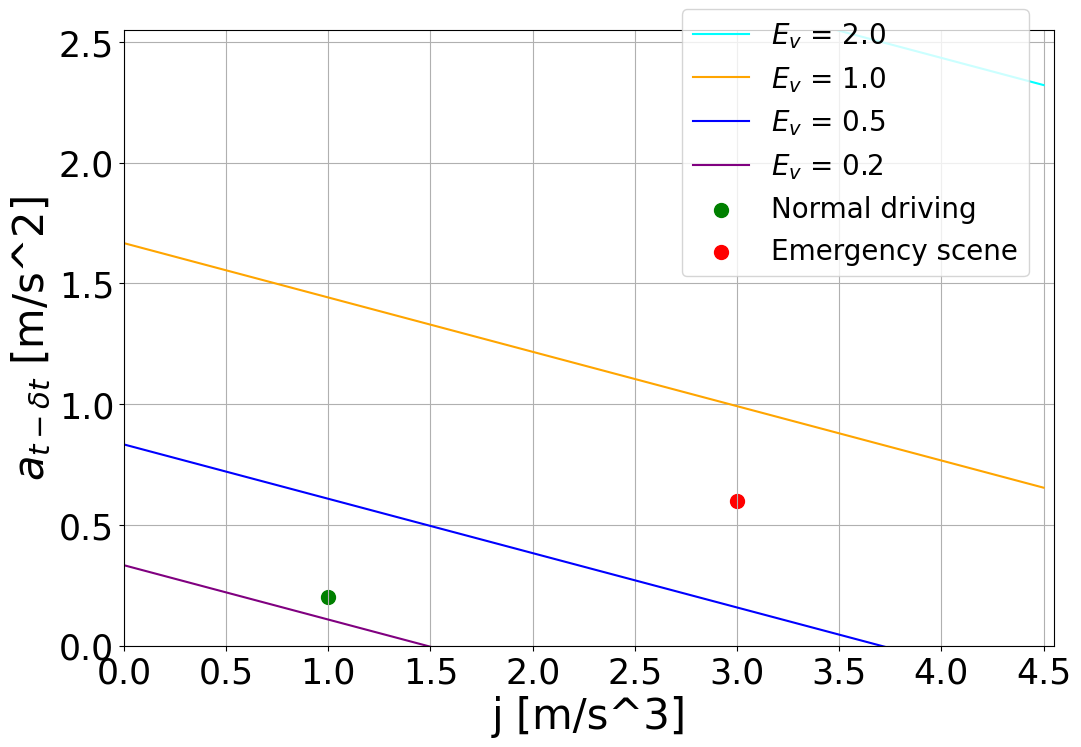}
      \hspace{1.6cm} (c) $\delta t = 0.2 $ (5 Hz annotation).
    \end{minipage} &
    \begin{minipage}[t]{0.40\hsize}
      \centering
      \includegraphics[width=0.9\linewidth]{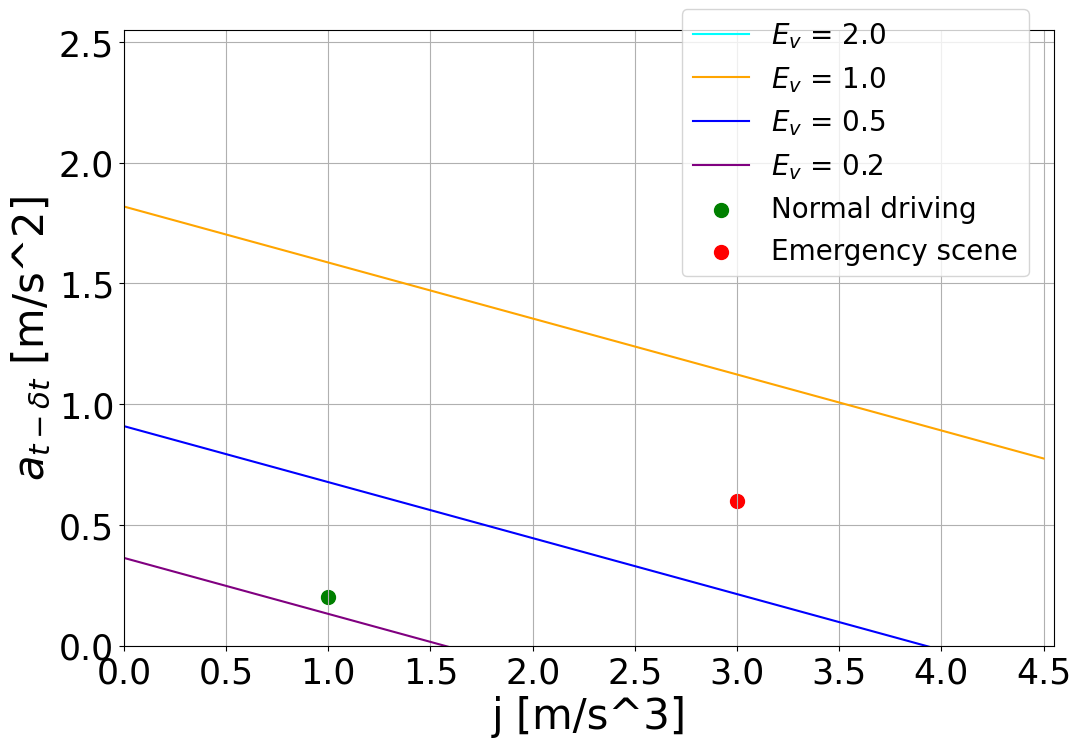}
      \hspace{1.6cm} (d) $\delta t = 0.1 $ (10 Hz annotation).
    \end{minipage}
  \end{tabular}
  \caption{
    The estimation error of $ v_{t + \Delta t}^{GT} $ for inference time $\Delta t$ = 0.5 s.
  }
  \label{v_DT05}
\end{figure*}

\begin{figure}[th]
    \begin{center}
        \includegraphics[width=0.9 \linewidth]{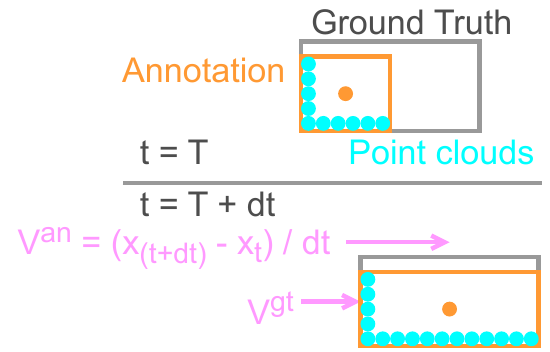}
        \caption{
          The error of annotation.
        }
        \label{annotation_error}
    \end{center}
\end{figure}

Here, we consider whether the annotation frequency is appropriate.
We assume the velocity of ego vehicle $ v^{ego}_t $ as \SI{0}{m/s}.
This assumption does not lose generality, as we can consider it in a relative coordinate system.

First, we analyze the estimation error of $ x_{t + \Delta t}^{GT} $.
The ground truth of the position $ x $ at the time $ (t + \Delta t) $, $ x_{t + \Delta t}^{*GT}$ is calculated by

\begin{equation}
    x_{t + \Delta t}^{*GT} = x_{t}^{GT} + v_{t}^{GT} \Delta t + \frac{1}{2} a_{t}^{GT} \Delta t^2 + \frac{1}{6} j \Delta t ^3
\end{equation}

, where $ a_{t}^{GT} $ is acceleration at the time $t$, $j$ is jerk.
In this work, The jerk is assumed to be constant.
The error of the position $E_x$ between $x_{t + \Delta t}^{*GT}$ and $ x_{t + \Delta t}^{GT} $ is calculated by

\begin{align}
    E_x &= x_{t + \Delta t}^{*GT} - x_{t + \Delta t}^{GT} \\
        &= \frac{(a_{t - \delta t}^{GT} \delta t + j \delta t ^2 )}{2} \Delta t + \frac{(a_{t - \delta t}^{GT} + j \delta t )}{2} \Delta t^2 + \frac{j}{6} \Delta t ^3.
  \label{e_x1}
\end{align}

\cref{e_x1} indicates that the estimation error decreases as $\delta t$ becomes smaller and $\Delta t$ becomes smaller.
Additionally, the estimation error decreases as $ a_{t - \delta t}^{GT} $ becomes smaller and $ j $ becomes smaller.

For each $\Delta t$ and $\delta t$, we visualized the relationship between estimation error $E_x$ = (\SI{0.05}{m}, \SI{0.1}{m}, \SI{0.2}{m}, \SI{0.5}{m}) and the corresponding values of $j$ and $ a_{t - \delta t}^{GT}$.
As reference values, we also plotted the cases of normal driving (j = \SI{1}{m/s^3}, $a_{t - \delta t}^{GT}$ = \SI{0.2}{m/s^2}) and emergency case (j = \SI{3}{m/s^3}, $ a_{t - \delta t}^{GT}$ = \SI{0.6}{m/s^2}).
\cref{x_DT02} shows the estimation error of $ x_{t + \Delta t}^{GT} $ for inference time $\Delta t$ = 0.2 s.
\cref{x_DT05} shows the estimation error of $ x_{t + \Delta t}^{GT} $ for inference time $\Delta t$ = 0.5 s.

In the nuScenes annotation, the annotation frequency is set as $ \delta t $ = \SI{0.5}{s} (2 Hz annotation).
When $ \Delta t $ = \SI{0.2}{s}, the estimation error is approximately \SI{0.05}{m} in a normal driving scenario.
This error is sufficiently small compared to commonly used thresholds (\SI{0.5}{m}, \SI{1.0}{m}, \SI{1.5}{m}, \SI{2.0}{m}), indicating that the annotation frequency is adequate for overall evaluation using mAP.
Even in emergency scenarios, the estimation error remains within \SI{0.2}{m}, which suggests that the overall evaluation is still valid.
However, when $ \Delta t $ = \SI{0.5}{s}, the estimation error increases to approximately \SI{0.2}{m} in normal driving scenarios.
In contrast, in emergency scenarios, the estimation error exceeds \SI{0.5}{m}, surpassing the threshold margin.
This indicates that an annotation interval of $ \delta t $ = \SI{0.5}{s} is insufficient.
To ensure an estimation error within \SI{0.2}{m}, as achieved when $\Delta t $ = \SI{0.2}{s},
a higher annotation frequency of $ \delta t $ = \SI{0.1}{s} (10 Hz annotation) is required.
This corresponds to a fivefold increase in annotation volume.
Thus, the real-time constraint $ \Delta t $ significantly impacts the annotation cost to maintain an acceptable level of estimation error.
Conversely, given the inference time $ \delta t $ and acceptable error in a system, the required annotation interval can be determined as a specification requirement.

Next, we analyze the estimation error of $ v_{t + \Delta t}^{GT} $.
The ground truth of the velocity $ v $ at the time $ (t + \Delta t) $, $ v_{t + \Delta t}^{*GT}$ is calculated by

\begin{equation}
    v_{t + \Delta t}^{*GT} = v_{t}^{GT} + a_{t}^{GT} \Delta t + \frac{1}{2} j \Delta t^2.
\end{equation}

The error of the velocity $E_y$ between $v_{t + \Delta t}^{*GT}$ and $ v_{t + \Delta t}^{GT} $ is calculated by

\begin{align}
  E_v &= v_{t + \Delta t}^{*GT} - \hat{v_{t}}^{GT} \\
      &= \frac{2 a_{t - \delta t}^{GT} \delta t + j \delta t ^2}{4} + a_{t}^{GT} \Delta t + \frac{j}{2} \Delta t^2
  \label{e_x2}
\end{align}

For each $\Delta t$ and $\delta t$, we visualized the relationship between estimation error $E_y$ = (\SI{0.2}{m/s}, \SI{0.5}{m/s}, \SI{1.0}{m/s}, \SI{2.0}{m/s}) and the corresponding values of $j$ and $ a_{t - \delta t}^{GT}$.

In the nuScenes annotation, for velocity estimation errors at $ \Delta t = \SI{0.2}{s} $, the error remains within \SI{0.2}{m/s} in normal scenarios and within \SI{1.0}{m/s} in emergency scenarios.
However, in emergency scenarios at $ \Delta t = \SI{0.5}{s} $, the velocity estimation error reaches approximately \SI{1.0}{m/s}, which may negatively impact downstream tracking and planning algorithms.
Additionally, if the velocity estimation accuracy of the model is comparable to this $E_v$, it becomes difficult to ensure a reliable evaluation of velocity estimation performance.
For instance, if $E_v < \SI{0.20}{m/s}$ and the mean velocity estimation error from model inference is \SI{0.3}{m/s}, it is reasonable to assume a velocity error of \SI{0.5}{m/s} and incorporate this assumption into the parameter settings of downstream tracking and planning modules.

Note that, in practical cases, annotation itself contains position errors.
As illustrated in \cref{annotation_error}, there are cases where only a part of a vehicle is visible in a given frame, and a 3D bounding box is annotated accordingly.
In the next annotated frame, the entire vehicle becomes visible, leading to a different annotation.
Since 3D bounding box annotations are based on LiDAR point clouds, occlusions or missing point clouds can result in incorrect annotations.
In such cases, velocity is calculated using the position difference between frames as:

\begin{equation}
    v^{an} = \frac{x^{an}_{t+\Delta t} - x^{an}_{t}}{\Delta t},
\end{equation}

where the annotation position error $x^{an}_{t}$ propagates into the velocity estimation $v^{an}$.
For instance like \cref{annotation_error}, if the annotation position error is \SI{0.5}{m} and the annotation interval is \SI{0.5}{s} (2 Hz annotation), the resulting velocity estimation error can reach \SI{1.0}{m/s}.
Therefore, it is crucial to take such annotation-induced errors into account when evaluating velocity estimation accuracy.

\end{document}